\documentclass{article}

\usepackage{PRIMEarxiv}

\usepackage[utf8]{inputenc} % allow utf-8 input
\usepackage[T1]{fontenc}    % use 8-bit T1 fonts
\usepackage{hyperref}       % hyperlinks
\usepackage{url}            % simple URL typesetting
\usepackage{booktabs}       % professional-quality tables
\usepackage{amsfonts}       % blackboard math symbols
\usepackage{nicefrac}       % compact symbols for 1/2, etc.
\usepackage{microtype}      % microtypography
\usepackage{lipsum}
\usepackage{fancyhdr}       % header
\usepackage{graphicx}       % graphics
\graphicspath{{media/}}     % organize your images and other figures under media/ folder
\usepackage{amsmath, amsthm, amssymb}

\usepackage{subcaption}

\usepackage{natbib}

\newtheorem{prop}{Proposition}
\newtheorem{cor}{Corollary}
\newtheorem{lemma}{Lemma}

\title{Inverting Self-Organizing Maps: A Unified Activation-Based Framework
}

\author{
  Alessandro Londei \\
  Sony Computer Science Laboratories - Rome\\
  Joint Initiative CREF-SONY, Centro Ricerche Enrico Fermi\\
  Via Panisperna 89/A, 00184\\
  Rome, Italy \\
  \texttt{alessandro.londei@sony.com} \\
  \And
  Matteo Benati \\
  Department of Computer, Automatic and Management Engineering\\
  Sapienza University, Via Ariosto 25\\
  Rome, Italy\\
  \texttt{matteo.benati@uniroma1.it} \\
   \And
  Denise Lanzieri \\
  Sony Computer Science Laboratories - Rome\\
  Joint Initiative CREF-SONY, Centro Ricerche Enrico Fermi\\
  Via Panisperna 89/A, 00184\\
  Rome, Italy \\
  \texttt{denise.lanzieri@sony.com} \\
  \AND
  Vittorio Loreto \\
  Sony Computer Science Laboratories - Rome\\
  Joint Initiative CREF-SONY, Centro Ricerche Enrico Fermi\\
  Via Panisperna 89/A, 00184\\
  Rome, Italy \\
  Physics Department\\
  Sapienza University, Piazzale Aldo Moro 1\\
  Rome, Italy\\
  Complexity Science Hub\\
  Josefstädter Strasse 39, A 1080, Vienna, Austria\\
  \texttt{vittorio.loreto@sony.com} \\
}

\begin{document}
\maketitle

\begin{abstract}
Self-Organizing Maps (SOMs) provide topology-preserving projections of 
high-dimensional data, yet their use as generative models remains 
largely unexplored.
We show that the activation pattern of a SOM---the squared distances to 
its prototypes---can be \emph{inverted} to recover the exact input, following from a classical result in Euclidean distance geometry: a  point in $D$ dimensions is uniquely determined by its distances to 
$D{+}1$ affinely independent references.
We derive the corresponding linear system and characterize the 
conditions under which inversion is well-posed.
Building on this mechanism, we introduce the \emph{Manifold-Aware 
Unified SOM Inversion and Control} (MUSIC) update rule, which modifies 
squared distances to selected prototypes while preserving others, 
producing controlled, semantically meaningful trajectories aligned with the SOM's piecewise-linear structure.  Tikhonov regularization 
stabilizes the update and ensures smooth motion in high dimensions.
Unlike variational or diffusion-based generative models, MUSIC requires 
no sampling, latent priors, or learned decoders: it operates entirely 
on prototype geometry.  If no perturbation is applied, inversion 
recovers the exact input; when a target prototype or cluster is 
specified, MUSIC produces coherent semantic transitions.
We validate the framework on synthetic Gaussian mixtures, MNIST digits, 
and the Labeled Faces in the Wild dataset.  Across all settings, MUSIC 
trajectories maintain high classifier confidence, produce significantly 
sharper intermediate images than linear interpolation, and reveal an interpretable geometric structure of the learned map.
\end{abstract}

% FIX: Replace placeholder keywords!
\keywords{Self-Organizing Maps \and Activation inversion \and 
Topology-preserving generation \and Latent space navigation \and 
Tikhonov regularization}

\section{Introduction}

Self--Organizing Maps (SOMs) \citep{kohonen1982self, kohonen2001basic} are 
classical topology--preserving neural models that embed high--dimensional data 
into a structured two-dimensional lattice of prototype vectors.  
Through competitive learning and neighbourhood cooperation, a SOM organizes 
prototypes so that nearby units respond to similar inputs, providing an 
interpretable low-dimensional representation of the data manifold.  
This property has made SOMs influential in applications ranging from 
image analysis \citep{hauske1997self} and speech processing 
\citep{kohonen2012self} to genomics \citep{nikkila2002analysis} 
and exploratory visualization \citep{vesanto2002data}.  
%EDIT: trimmed from 6 domains to 4; remote sensing and robotics cut
Over the years, several architectural variants---such as Growing Neural Gas 
\citep{fritzke1995growing}, Dynamic SOMs \citep{alahakoon2000dynamic}, 
hierarchical maps \citep{dittenbach2000growing}, and prototype--regularized 
deep models like SOM--VAE \citep{fortuin2018som}---have extended the 
expressive power of SOM--based learning while preserving its foundation in 
prototype geometry.

Traditionally, SOMs are used as a form of structured vector quantization:  
each input $z \in \mathbb{R}^D$ is associated with its Best--Matching Unit 
(BMU), and the BMU index serves as a discrete representation of the data.  
In this work, we adopt a fundamentally different viewpoint.  
Rather than mapping data \emph{to} prototypes, we show that the SOM 
activation pattern---specifically, the squared distances from $z$ to all 
prototypes---contains enough geometric information to \emph{reconstruct} $z$ 
exactly (Section~\ref{sec:theory}).
%EDIT: deleted three redundant sentences that repeated this claim;
%      added short forward-ref "(Section~X)" instead.

Building on this observation, we introduce a unified geometric framework for 
continuous semantic transformations in the data space.  
Our method proceeds in three steps:

\begin{enumerate}
    \item \textbf{Exact inversion.}  
    From the squared--distance activations to the prototypes, solve a linear 
    system to recover the input vector $z$.  When the SOM prototypes span 
    $\mathbb{R}^D$, inversion is exact up to machine precision.

    \item \textbf{Structured perturbation of activations.}  
    Selectively modify the squared distances to a chosen subset of prototypes 
    (e.g., a target cluster or a single prototype) while preserving the 
    others.  These perturbations must satisfy Euclidean realizability 
    constraints, which we enforce via a Tikhonov--regularized linear system.

    \item \textbf{MUSIC update rule.}  
    The resulting \emph{Manifold--Aware Unified SOM Inversion and Control} 
    (MUSIC) update produces a latent step $\Delta z$ that simultaneously 
    moves toward the desired prototypes while maintaining consistency with 
    the preserved distances.
    %EDIT: "Manifold-Aware" → "Topology-Aware"
\end{enumerate}

This framework turns a SOM from a static discretizer into a continuous, 
interpretable mechanism for latent--space exploration.  
In contrast to generative models that rely on learned decoders and latent 
priors \citep{kingma2013auto, van2017neural, ho2020denoising}, MUSIC 
operates entirely on prototype geometry.  
If no conditioning is applied, inversion reproduces the exact input; when 
conditioning is introduced, MUSIC yields semantically meaningful variations 
that remain on the manifold encoded by the SOM.

We validate the method across datasets of increasing complexity: synthetic 
Gaussian mixtures (where ground truth geometry is known), handwritten digits 
(MNIST), and natural face images (Labeled Faces in the Wild).  
Across all settings, MUSIC generates coherent semantic trajectories that 
maintain high classifier confidence, produce significantly sharper 
intermediate images than linear interpolation, and preserve local identity 
where appropriate.

The contributions of this paper are therefore twofold:  
(i)~showing that classical SOM activations admit exact inversion through 
distance geometry, and 
(ii)~introducing a principled update rule that turns the SOM into a 
geometry--aware engine for semantic latent--space navigation.

%%%%%%%%%%%%%%%%%%%%%%%%%%%%%%%%%%

\section{Inverting Self--Organizing Maps}
\label{sec:theory}

Let $\{w_j\}_{j=1}^N \subset \mathbb{R}^D$ denote the prototype (weight) 
vectors of a trained SOM.  For any input vector $z \in \mathbb{R}^D$, 
define the \emph{squared-distance activation vector}
\[
a(z) \;=\;
\begin{bmatrix}
a_1(z)\\[-2pt]
\vdots\\[-2pt]
a_N(z)
\end{bmatrix},
\qquad 
a_j(z) \;=\; \|z - w_j\|^2.
\]
We work with \emph{squared} distances because $a_j(z)$ has linear 
first--order variation, 
$\mathrm{d}a_j = 2(z-w_j)^\top \mathrm{d}z$, yielding a well--conditioned 
local Jacobian.
Moreover, subtracting any reference activation $a_r(z)$ removes the common 
quadratic term $\|z\|^2$, exposing an \emph{affine} dependence on $z$ that 
is crucial for exact inversion.
The mapping $f\!:\mathbb{R}^D \!\to\! \mathbb{R}^N$, $f(z)=a(z)$, can be 
interpreted as the activation state of the SOM when exposed to~$z$.  
Recovering $z$ from its distances to a sufficiently rich set of references 
is a standard consequence of Euclidean distance geometry and multilateration 
\citep{blumenthal1953distance, gower1985properties, liberti2014euclidean, 
meyer2021solving}.
%EDIT: equivariance paragraph moved to Supplementary S1
The activations are moreover invariant under rigid motions of the 
prototype set (Appendix~\ref{supp:equivariance}), so only the relative 
configuration of prototypes matters for inversion.

\subsection{Exact Inversion Property}
\label{subsec:theory_inversion}

The following proposition establishes that $f$ is injective and exactly 
invertible under a mild rank condition on the prototypes.

\begin{prop}[Exact Inversion Property]
\label{prop:exact_inverse}
If the prototype set $\{w_j\}_{j=1}^N$ spans $\mathbb{R}^D$, 
i.e.\ $\mathrm{rank}([w_1 - w_N,\, \dots,\, w_{N-1} - w_N]) = D$,  
then the mapping $f$ is injective.  
Moreover, the input vector $z$ can be exactly reconstructed from the 
activation differences $\{a_j(z) - a_N(z)\}_{j=1}^{N-1}$ by solving a 
linear system:
\[
z \;=\; (B^\top B)^{-1} B^\top c,
\]
where
\[
B \;=\; 
\begin{bmatrix}
2(w_N - w_1)^\top \\[-2pt]
\vdots \\[-2pt]
2(w_N - w_{N-1})^\top
\end{bmatrix}
\in\mathbb{R}^{(N-1)\times D}, 
\qquad
c \;=\;
\begin{bmatrix}
a_1(z) - a_N(z) + \|w_N\|^2 - \|w_1\|^2 \\[-2pt]
\vdots \\[-2pt]
a_{N-1}(z) - a_N(z) + \|w_N\|^2 - \|w_{N-1}\|^2
\end{bmatrix}.
\]
\end{prop}

\begin{proof}
Each squared activation expands as
$a_j(z) = z^\top z - 2 z^\top w_j + \|w_j\|^2$.
Subtracting $a_N(z)$ eliminates the quadratic term:
\[
a_j(z) - a_N(z) 
\;=\;
2(w_N - w_j)^\top z
+ \|w_j\|^2 - \|w_N\|^2, 
\qquad j = 1, \dots, N-1.
\]
This defines an affine system $Bz = c$ with $B$ and $c$ as above.  
If $B$ has full column rank~$D$, the Moore--Penrose pseudoinverse gives
\begin{equation}
z \;=\; B^{+} c \;=\; (B^\top B)^{-1} B^\top c,
\end{equation}
and the reconstruction is exact.
\end{proof}

The reference index~$N$ is arbitrary: any anchor $r\in\{1,\dots,N\}$ yields 
an equivalent system.  In practice, centering the prototypes and optionally 
whitening can improve the conditioning of~$B$.
This linearization connects to classical Euclidean distance geometry: 
activation differences form an affine slice of a Euclidean Distance Matrix 
(EDM), whose embedding is unique up to rigid motions under mild rank 
conditions 
\citep{blumenthal1953distance, gower1985properties, asimow1978rigidity}.

\subsection{Stability and Conditioning}

\begin{cor}[Stability to activation noise]
\label{cor:stability}
Let $\tilde a_j(z)=a_j(z)+\varepsilon_j$ with 
$\varepsilon\sim\mathcal{N}(0,\sigma^2 I)$.  If $B$ has full column rank, 
the least--squares inverse satisfies
\[
\|\hat z - z\| \;\le\; \|B^+\|\,\sigma\,\sqrt{N-1}.
\]
Hence, inversion is Lipschitz--stable, with amplification governed by the 
inverse of the smallest singular value of~$B$.
\end{cor}

\begin{figure}[t]
    \centering
    \includegraphics[scale=0.75]{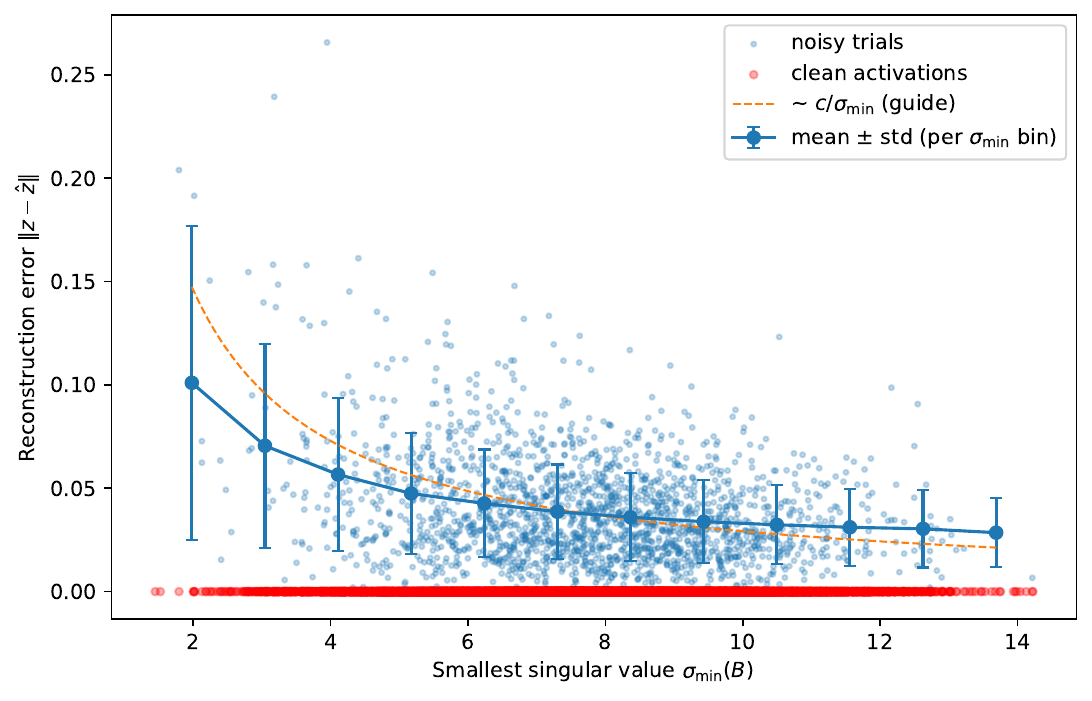}
    \caption{\textbf{Inversion accuracy vs.\ conditioning of the prototype 
    geometry.}
    Clean activations (red) yield reconstructions at machine precision.
    Adding Gaussian noise induces errors that grow as 
    $1/\sigma_{\min}(B)$: well-spread prototypes give stable inversion, 
    while nearly collinear configurations amplify noise.
    Blue circles show binned means $\pm$ standard deviations over 
    $20{,}000$ trials.}
    \label{fig:inversion_accuracy}
\end{figure}

Figure~\ref{fig:inversion_accuracy} empirically validates this bound: 
across $20{,}000$ noisy trials, the reconstruction error 
$\|z-\hat z\|_2$ grows as $\sigma_{\min}(B)$ decreases, following the 
predicted scaling
\[
\mathbb{E}\,\| \hat z - z \|_2^2 
\;=\; \sigma^2\,\mathrm{tr}\!\big((B^\top B)^{-1}\big)
\;\lesssim\; \sigma^2\,\frac{D}{\sigma_{\min}^2(B)}\,.
\]

For overcomplete maps ($N > D{+}1$), the system is consistent and 
inversion reduces to least squares, with reconstruction errors at 
double-precision level (${\sim}10^{-18}$).
A weighted least-squares extension for correlated noise is given in 
Appendix~\ref{supp:weighted_ls}.

When the ambient dimension exceeds the rank of the prototype matrix 
(e.g., raw pixel inputs), dimensionality reduction (PCA or an 
autoencoder bottleneck) is necessary before inversion can succeed.
%EDIT: Lemma 1 (rank ↔ affine independence) cut—restates the 
%      Proposition's hypothesis.
The Jacobian $J(z)\in\mathbb{R}^{N\times D}$ with rows 
$J_j(z) = 2(z-w_j)^\top$ has full rank~$D$ for all 
$z\in\mathbb{R}^D$ whenever the prototypes contain $D{+}1$ affinely 
independent points (Appendix~\ref{supp:jacobian_proof}); we use this 
in the perturbation analysis of Section~\ref{sec:PerturbingSOM}.
%EDIT: Lemma 2 statement kept inline; proof moved to Supplementary.

This exact-inversion property establishes the SOM activation map as a 
mathematically well-posed latent representation.
In the next section, we exploit it to define small, consistent 
perturbations in activation space that translate into interpretable 
input--space variations.

%%%%%%%%%%%%%%%%%%%%%%%%%%%%%%%%%%

\section{Perturbing the SOM activation layer}
\label{sec:PerturbingSOM}

Having established that the activation map admits exact inversion, we 
now use it as a control interface.  We perturb distances to selected 
prototypes and recover the corresponding latent update by solving a 
regularized inverse problem, yielding smooth and semantically meaningful 
variations of~$z$.

\begin{figure}[t]
    \centering
    \includegraphics[scale=0.9]{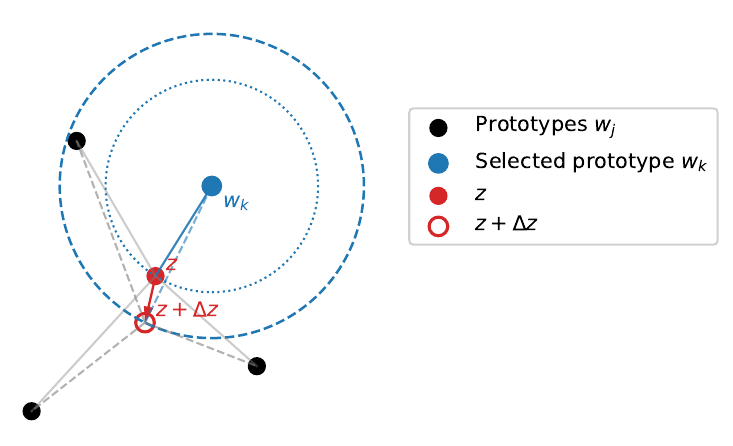}
    \caption{\textbf{Geometric schematic of a single--cell perturbation in 
    the SOM activation space.}  A small displacement $\Delta z$ of the 
    input vector $z$ modifies its distance to the selected prototype 
    $w_k$, thereby inducing a consistent reconfiguration of distances to 
    all other prototypes $w_j$.  Solid lines indicate the original 
    distances $\|z-w_j\|$, and dashed lines the updated distances 
    $\|z+\Delta z-w_j\|$.  Because all distances are mutually constrained 
    by the geometry of $\mathbb{R}^D$, a perturbation of one activation 
    cell necessarily entails a coordinated adaptation of the others to 
    maintain a valid Euclidean configuration.}
   \label{fig:som_perturbation_schematic}
\end{figure}

We denote by $\mathrm{BMU}(z)=\arg\min_j \|z-w_j\|$ the best--matching 
unit and by $\mathcal{V}_j$ its Voronoi region \citep{58325}.
The collection $\{\mathcal{V}_j\}$ forms a Voronoi tessellation of the 
input space, providing the piecewise-linear partition underlying the SOM 
topology.
Each input $z \in \mathbb{R}^D$ induces an activation pattern 
$a(z)=\{\|z-w_j\|^2\}_{j=1}^N$; small, geometrically consistent 
perturbations of this pattern yield coherent variations of~$z$, provided 
the modified activations remain compatible with a valid Euclidean 
configuration.
Arbitrary changes of individual activations generally violate this 
consistency, as all prototype distances are mutually dependent
\citep{58325, cottrell1998theoretical, vesanto1999som}.
Figure~\ref{fig:som_perturbation_schematic} illustrates this geometric
interdependence for a perturbation applied to a single prototype.

\vspace{4pt}
\noindent
To introduce a valid perturbation, we seek the smallest change
$\Delta z\!\in\!\mathbb{R}^D$ that produces a prescribed modification in 
the distance to a selected prototype while approximately preserving the 
geometry of the other distances.
Let
\[
z\in\mathbb{R}^D,\qquad
\{w_j\}_{j=1}^N\subset\mathbb{R}^D,\qquad
k\in\{1,\dots,N\},\qquad
\Delta d_k\in\mathbb{R},
\]
where $w_j$ are the prototype vectors and $\Delta d_k$ is the desired
increment in the distance between $z$ and $w_k$.
The goal is to determine a perturbation vector $\Delta z$ satisfying:
\begin{enumerate}
    \item the distance to prototype $k$ changes by $\Delta d_k$;
    \item the norm $\|\Delta z\|$ is minimal (scaled by a user--defined 
          parameter $r_{\text{scale}}$);
    \item the relative geometry of all other distances is approximately 
          preserved.
\end{enumerate}

Defining the relative position and current distance
\[
x := z - w_k, \qquad d_k := \|x\|, \qquad d_k' := d_k + \Delta d_k,
\]
the desired squared change in distance is
\[
\Delta d_k^2 = (d_k')^2 - d_k^2 = \|x+\Delta z\|^2 - \|x\|^2.
\]
Expanding gives
\[
\|x+\Delta z\|^2 = \|x\|^2 + 2x^\top\Delta z + \|\Delta z\|^2,
\]
and thus
\[
2x^\top\Delta z + \|\Delta z\|^2 = \Delta d_k^2.
\]
For infinitesimal perturbations, the quadratic term is second--order 
and can be neglected, yielding the first--order constraint:
\[
x^\top\Delta z = c := \tfrac{1}{2}\Delta d_k^2.
\]
This fixes the projection of $\Delta z$ along~$x$.
Decomposing the perturbation as
\[
\Delta z = \Delta z_{\parallel} + \Delta z_{\perp},\qquad
\Delta z_{\parallel} = \frac{c}{\|x\|^2}x,\qquad
x^\top\Delta z_{\perp}=0,
\]
one obtains a parallel component that achieves the desired change in 
distance and an orthogonal component spanning the local isodistance 
subspace.
Geometrically, $\Delta z_{\perp}$ belongs to the tangent space of the 
SOM distance manifold: in the linear regime, such directions leave all squared distances approximately unchanged to first order.
Crucially, in high dimensions, tangential perturbations become 
increasingly close to true isometries: the first-order distance 
variations satisfy $|\Delta a_j| \sim \|\Delta z_\perp\|\,\|c_j\|/\sqrt{D}$ 
(where $c_j = w_k - w_j$), so that the relative distortion of 
non-target distances vanishes as $D$ grows, while the 
$O(\|\Delta z\|^2)$ remainder of the Taylor expansion controls the 
ultimate accuracy of the linear approximation.
A formal derivation of this concentration effect is given in 
Supplementary Section~\ref{supp:S1}.

Importantly, the accuracy of this linearisation can be characterised
exactly for squared-distance activations.  Since each
$a_j(z)=\|z-c_j\|^2$ is quadratic in~$z$, its Taylor expansion
terminates at second order:
\[
a_j(z+\Delta z) \;=\; a_j(z) \;-\; 2(c_j-z)^\top\Delta z
\;+\; \|\Delta z\|^2 .
\]
The second-order residual $\|\Delta z\|^2$ is independent of~$j$:
it shifts all activations by the same amount.  In the anchored system
used throughout this work, which operates on activation
\emph{differences} $a_j - a_r$ relative to a reference prototype~$r$,
this common term cancels exactly, leaving
\[
(a_j - a_r)(z{+}\Delta z) - (a_j - a_r)(z)
\;=\; -2(c_j - c_r)^\top \Delta z ,
\]
which is linear in $\Delta z$ with a constant coefficient matrix.
The first-order Jacobian model is therefore not an approximation
but the exact relationship for activation differences under
squared-distance activations.  In absolute terms, the sole residual
is the uniform bias $\|\Delta z\|^2 = O(\rho^2)$, which does not
affect the direction of the optimal update, only its magnitude.

In absolute (non-anchored) terms, the sole residual is the
uniform bias $\|\Delta z\|^2$, which is $O(\rho^2)$ under the
step-size clipping enforced in MUSIC and does not affect the
\emph{direction} of the optimal update $\Delta z$, only its
magnitude.
This structural property of quadratic activations is what
distinguishes the MUSIC linearisation from generic Taylor
approximations, where both the magnitude and direction of the
remainder depend on the nonlinear form of the activation function
and can accumulate unpredictably.

\begin{figure}[t]
    \centering
    \includegraphics[scale=0.9]{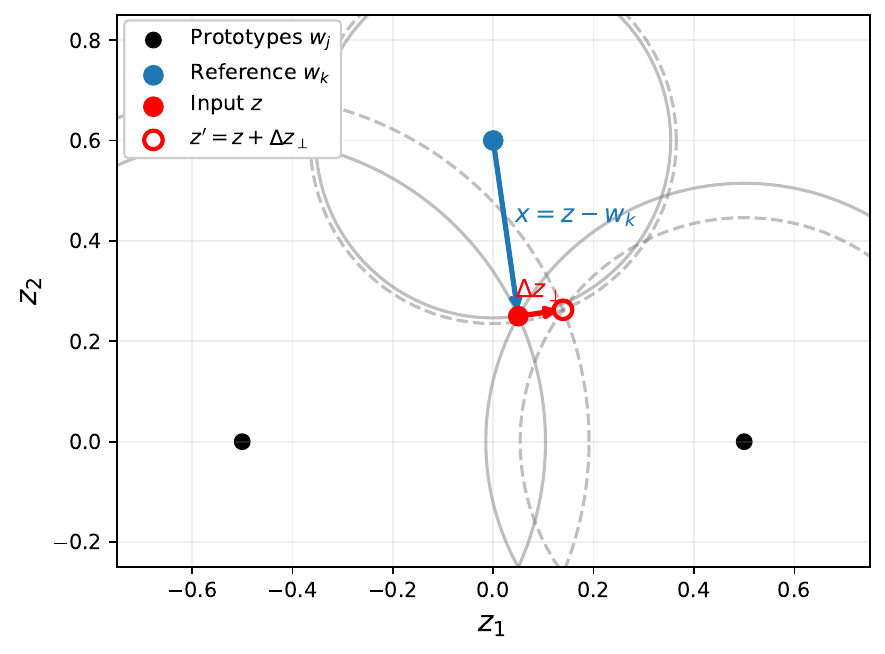}
    \caption{\textbf{Radial-tangential decomposition around a reference 
    prototype.}
Black dots denote the SOM prototypes $w_j$, and the blue point marks 
the reference prototype $w_k$. 
The red point shows the current input $z$, while the unfilled red point 
indicates the tangentially displaced location $z' = z + \Delta z_{\perp}$.
The blue arrow represents the radial direction $x = z - w_k$, which 
determines the first-order change of the squared distance to the 
reference prototype. 
The red arrow shows a tangential perturbation $\Delta z_{\perp}$ 
orthogonal to $x$ at $z$. 
For each prototype, the solid circle traces the locus of points at the 
same distance as $z$, while the dashed circle shows the corresponding 
locus for $z'$. 
The circles centered at $w_k$ coincide at first order, and the circles 
centered at the other prototypes remain nearly unchanged in the 
neighborhood of $z$, illustrating that tangential motion approximately 
preserves all squared distances simultaneously.}
   \label{fig:perturbation_orthogonal_component}
\end{figure}

A geometric illustration of the radial--tangential decomposition and 
its effect on all SOM prototypes is provided in 
Figure~\ref{fig:perturbation_orthogonal_component}.

To control amplitude, we set a total perturbation norm
$r := r_{\text{scale}}\|\Delta z_{\parallel}\|$ and assign
\[
\|\Delta z_{\perp}\| = \sqrt{r^2 - \|\Delta z_{\parallel}\|^2}.
\]
The resulting perturbation combines a controlled radial shift with a 
tangential adjustment that minimally disturbs the remaining distances.

The image $\mathcal{A}=\{a(z):z\in\mathbb{R}^D\}\subset\mathbb{R}^N$ 
forms a smooth $D$--dimensional submanifold of activation space.
By the full--rank Jacobian property 
(Appendix~\ref{supp:jacobian_proof}), the tangent space at each point 
is $T_{a(z)}\mathcal{A}=\mathrm{Im}\,J(z)$ with 
$J_j(z)=2(z-w_j)^\top$.
Any activation vector $\tilde a\notin \mathcal{A}$ violates the 
consistency relations
\[
a_i(z)-a_j(z)=2(w_j-w_i)^\top z+\|w_i\|^2-\|w_j\|^2
\]
and cannot correspond to a valid point in $\mathbb{R}^D$.
Perturbing a single activation in isolation therefore moves $a(z)$ 
off~$\mathcal{A}$; to recover a consistent state, the modified 
activation must be reprojected by solving
\[
\hat z = \arg\min_{z\in\mathbb{R}^D}\|a(z)-\tilde a\|^2.
\]
This motivates the MUSIC framework (Section~\ref{sec:MUSIC}), which 
casts multi-prototype activation perturbations as the solution of a 
Tikhonov-regularized inverse problem balancing target attraction and 
global geometry preservation.

%%%%%%%%%%%%%%%%%%%%%%%%%%%%%%%%%%

\section{The MUSIC framework}
\label{sec:MUSIC}

Building on the geometric constraints of the previous section, we now 
seek a principled way to steer the SOM activation state while preserving 
the consistency of its distance representation.
A single-cell perturbation (Section~\ref{sec:PerturbingSOM}) can be 
analytically described, but extending this to multiple prototypes or continuous trajectories requires a unified formulation 
\citep{kiviluoto1996topology}.

We introduce the \emph{Manifold-Aware Unified SOM Inversion and 
Control} (MUSIC) framework, in which all activation-based perturbations 
are expressed as Tikhonov-regularized inverse problems.
Given a desired change in one or more prototype activations, MUSIC 
computes the minimal, geometrically consistent perturbation of the 
input vector, balancing three objectives:
(i)~preservation of non-target activations,
(ii)~attraction toward the desired activation changes, and
(iii)~regularization for smoothness and numerical stability.
The resulting update admits a closed-form solution, interpretable both 
as a weighted projection in input space and as the MAP estimate of a 
Gaussian model (Appendix~\ref{supp:S4}).

\subsection{Energy formulation and update rule}

MUSIC computes at each iteration a perturbation 
$\Delta z\in\mathbb{R}^D$ that realizes a desired change in 
squared-distance activations while preserving the remaining geometry.  
Let $a_j(z)=\|z-w_j\|^2$ with Jacobian row 
$J_j(z) = 2\,(z-w_j)^\top$.
Given a preserved index set~$S$ and a target set~$T$, we form $A_S$ 
and $B_T$ by stacking the (optionally row-normalized) Jacobian rows 
$\{\widehat{J}_j\}_{j\in S}$ and $\{\widehat{J}_t\}_{t\in T}$, 
evaluated at the current iterate~$z_t$ 
(see Appendix~\ref{supp:S5} for row-normalization details).

MUSIC chooses $\Delta z$ as the minimizer of the Tikhonov-regularized 
energy (Appendix~\ref{supp:S2} derives the full functional):
\begin{equation}
E_\gamma(\Delta z)
\;=\;
(1-\gamma)\,\|A_S\Delta z\|^2
\;+\;
\gamma\,\|B_T\Delta z - b\|^2
\;+\;
\lambda\,\|\Delta z\|^2,
\qquad \gamma\in[0,1],\ \lambda>0,
\label{eq:MUSIC_energy}
\end{equation}
where $b$ encodes the desired first-order target changes
(e.g., $b_t = \tfrac{1}{2}\Delta a_t$).  
The normal equations are
\begin{equation}
\Big[(1-\gamma)A_S^\top A_S + \gamma B_T^\top B_T + \lambda I\Big]
\,\Delta z^\star
\;=\;
\gamma\,B_T^\top b.
\label{eq:MUSIC_normal}
\end{equation}
With $H = (1-\gamma)A_S^\top A_S + \gamma B_T^\top B_T + \lambda I 
\succ 0$ (positive definite for any $\lambda>0$), the unique minimizer 
is $\Delta z^\star = H^{-1}\gamma\,B_T^\top b$.
Geometrically, $\lambda\|\Delta z\|^2$ induces an isotropic trust 
region in the tangent space, damping motion in poorly supported 
directions.
In practice we enforce a step constraint 
$\|\Delta z^\star\|\le \rho$ and relinearize $A_S, B_T$ at 
$z_{t+1}=z_t+\Delta z^\star$ to maintain first-order accuracy.

The Tikhonov parameter~$\lambda$ acts as a spectral low-pass filter on 
the combined constraint system: directions with small singular values 
(ill-conditioned or conflicting constraints) are attenuated, while 
well-supported modes pass nearly unaltered.  A detailed spectral 
analysis with filtering bounds is given in Appendix~\ref{supp:S3}.

Iterating the update yields a discrete trajectory that, in the 
small-step limit, follows the regularized gradient flow 
$\dot z = -\nabla_z E_\gamma(z)$, connecting the inverse-problem 
formulation to Riemannian gradient dynamics on the SOM distance manifold
\citep{absil2008optimization, amari2016information}.

In practice, MUSIC does not enforce exact isometries.  Each step 
balances modifying the target distances with approximately preserving 
the complement, using the tangent-space geometry of 
Section~\ref{sec:PerturbingSOM} to bias the evolution toward 
manifold-consistent, semantically meaningful changes.
The preservation set~$S$ may include all non-targets or an $r$-ring 
lattice neighborhood around the current BMU.
Empirically stable trajectories are obtained with 
$\gamma\in[0.7,0.95]$, $\lambda$ chosen via the L-curve criterion 
\citep{hansen1998rank}, and a trust radius 
$\rho = 0.02\,\|z-w_{\mathrm{BMU}}\|$.

\subsection{Exploration modes: free, informed, and cluster}
\label{subsec:MUSIC_exploration_modes}

The MUSIC energy~\eqref{eq:MUSIC_energy} specializes to three regimes 
by choosing the sets $(S, T)$ and the target vector~$b$.
In all cases we work with the linearized operators at~$z_t$:
\[
J_j(z_t)=2\,(z_t-w_j)^\top,\qquad
A_S \!=\! \text{stack}\{\,\widehat J_j: j\in S\,\},\quad
B_T \!=\! \text{stack}\{\,\widehat J_t: t\in T\,\},
\]
with optional diagonal weights $W_S, W_T$ to emphasize local topology.
Step-by-step pseudo-code for all three modes, including noise injection 
for enhanced coverage, is given in Appendix~\ref{supp:algorithms}.

\subsubsection*{(i) Free exploration (no explicit target)}
\label{subsubsec:free_exploration}

Free exploration seeks geometry-preserving motion without attraction 
toward any prototype: a local random walk on the activation manifold.
Removing the target term in~\eqref{eq:MUSIC_energy} 
($B_T=0$, $b$ omitted) yields
\begin{equation}
\min_{\Delta z}\ \|W_S A_S \Delta z\|^2 \;+\; \lambda \|\Delta z\|^2
\quad\text{s.t.}\quad \|\Delta z\|\le \tau .
\label{eq:free_explore}
\end{equation}
The trivial minimizer is $\Delta z=0$; one therefore selects a 
direction in which preserved activations change minimally.
Let $C = (W_S A_S)^\top (W_S A_S) + \lambda I$.
Any unit eigenvector $q_{\min}$ of the smallest eigenvalue of~$C$ 
identifies the least disruptive direction, giving 
$\Delta z^\star = \tau\, q_{\min}$.
Stochastic exploration is obtained by sampling from the low-eigenvalue 
subspace of~$C$, optionally adding a small radial component toward 
the current BMU to avoid drift.

\subsubsection*{(ii) Informed exploration (single target)}
\label{subsubsec:informed_single}

Informed exploration moves the input toward a chosen prototype~$t$ 
while preserving non-target distances.
Let $T=\{t\}$, $S=\{1,\dots,N\}\setminus\{t\}$, and set
\[
B_T = \widehat J_t,\qquad
b = -\,\eta\, a_t(z_t)\quad (\eta\in(0,1]),
\]
requesting a fractional reduction of the target activation.  
The MUSIC step solves
\begin{equation}
\min_{\Delta z}\ \|W_S A_S \Delta z\|^2 \;+\; \gamma\,\|W_T (B_T \Delta z - b)\|^2 \;+\; \lambda\|\Delta z\|^2.
\label{eq:single_target_quadratic}
\end{equation}
This single-target variant recovers the radial/tangential 
decomposition of Section~\ref{sec:PerturbingSOM} in the limit of 
vanishing~$\lambda$, while providing a numerically stable, one-shot 
update integrating preservation and attraction under a single convex 
objective.

\subsubsection*{(iii) Cluster exploration (multi-target)}
\label{subsubsec:cluster_multi}

Cluster exploration generalizes to coordinated motion toward a group 
of related prototypes $T$, preserving distances to the complement 
$S=\{1,\dots,N\}\setminus T$.  We solve
\begin{equation}
\min_{\Delta z}\ \underbrace{\|W_S A_S \Delta z\|^2}_{\text{preserve non-targets}}
\;+\;\gamma\,\underbrace{\|W_T (B_T \Delta z - b)\|^2}_{\text{shrink targets}}
\;+\;\lambda\|\Delta z\|^2,
\label{eq:multi_target_energy}
\end{equation}
with $b_t=-\eta\,a_t(z_t)$ for each $t\in T$.
Writing $A_w=W_S A_S$, $B_w=\sqrt{\gamma}\,W_T B_T$, and 
$b_w=\sqrt{\gamma}\,W_T b$, the normal equations are
\begin{equation}
\big(A_w^\top A_w + B_w^\top B_w + \lambda I\big)
\,\Delta z^\star \;=\; B_w^\top b_w.
\label{eq:multi_target_normal}
\end{equation}
We update $z_{t+1}=z_t+\Delta z^\star$, enforce the trust radius, 
and relinearize; $m=1$--$3$ passes typically suffice.
When $T$ is large, Gaussian weights in $W_T$ emphasize nearby 
targets and improve conditioning.

A lighter variant, \emph{random informed-within-cluster} exploration, 
draws a single prototype $t^{(i)}\sim\mathrm{Uniform}(T)$ at each 
step and applies the informed update toward $t^{(i)}$ alone.
In expectation this approximates the full multi-target control while 
reducing per-step cost, and empirically improves mixing by avoiding 
bias toward the cluster barycenter.

\begin{figure}[t]
  \centering
  \begin{subfigure}[t]{0.32\textwidth}
    \centering
    \includegraphics[width=\linewidth]{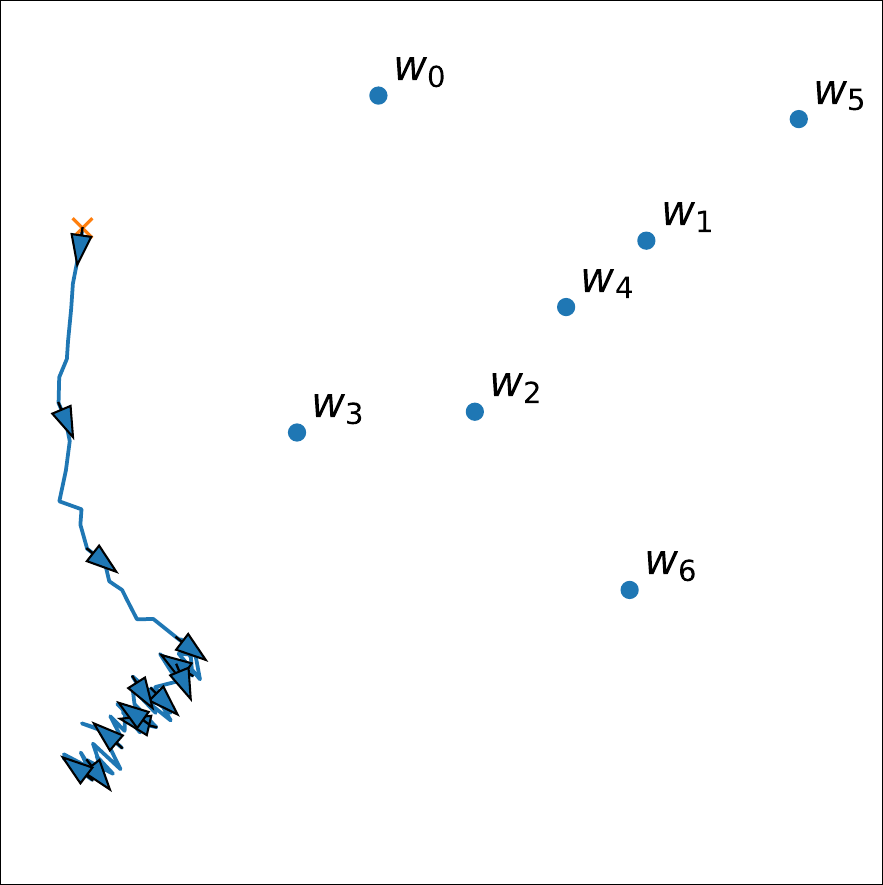}
    \caption{Free exploration.}
    \label{fig:exploration-free}
  \end{subfigure}\hfill
  \begin{subfigure}[t]{0.32\textwidth}
    \centering
    \includegraphics[width=\linewidth]{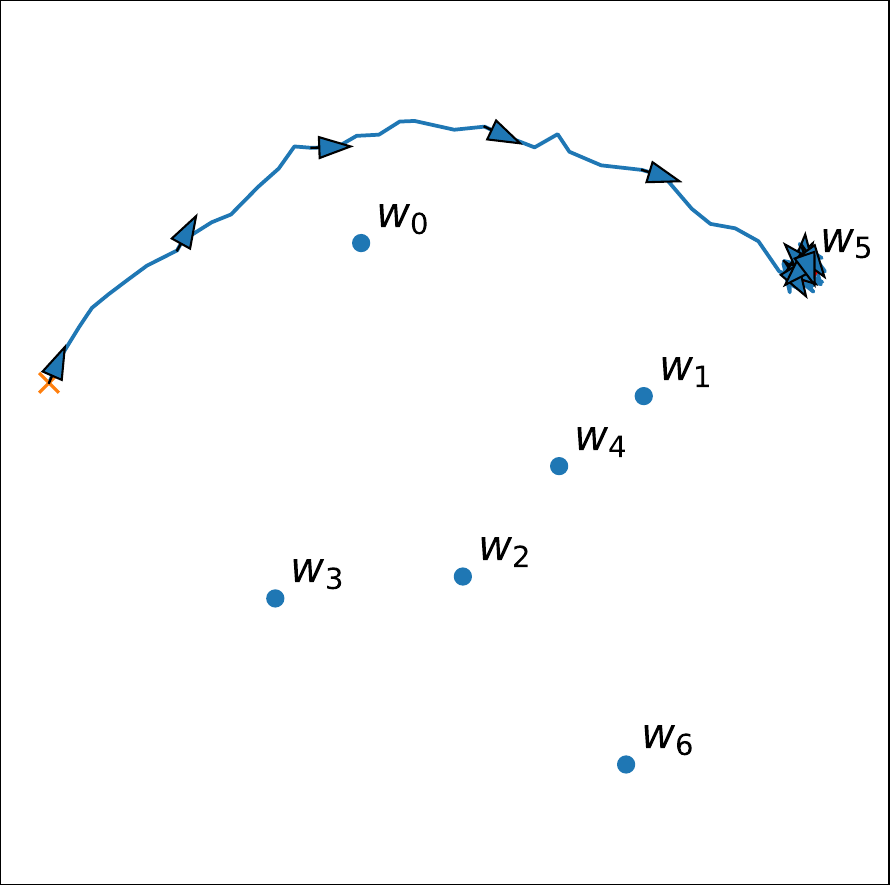}
    \caption{Informed exploration.}
    \label{fig:exploration-informed}
  \end{subfigure}\hfill
  \begin{subfigure}[t]{0.32\textwidth}
    \centering
    \includegraphics[width=\linewidth]{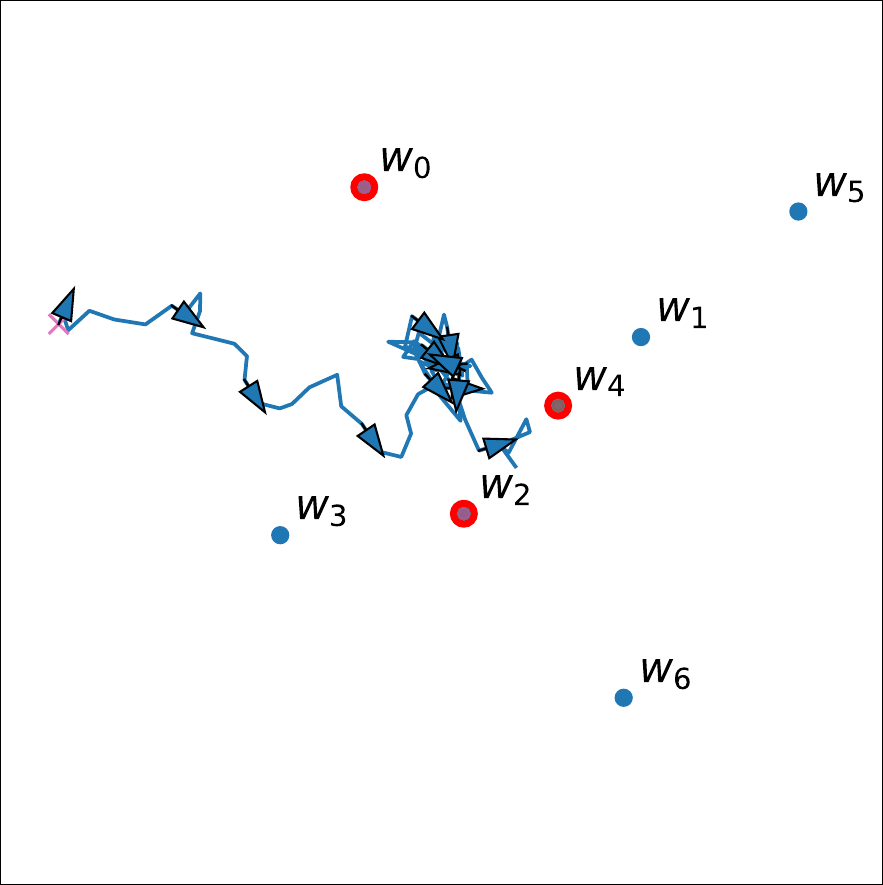}
    \caption{Cluster exploration.}
    \label{fig:exploration-cluster}
  \end{subfigure}
  \caption[Exploration modes on a SOM activation landscape]{%
  \textbf{Exploration modes with consistent activation editing (2D toy).}
  Seven prototypes (dots) are placed in $[-2.5,2.5]^2$; all 
  trajectories start from the same $z_0$ (cross).
  \emph{(a)}~Free: the update follows the least-perturbing eigenvector 
  of $J(z)^\top J(z)$, producing gentle drift.
  \emph{(b)}~Informed: a single target $w_t$ (hollow circle) is pulled 
  closer via the Tikhonov step while preserving other distances.
  \emph{(c)}~Cluster: a target set $T$ (hollow red circles) is 
  attracted with random sub-sampling per step, encouraging smooth 
  blending across the cluster region.}
  \label{fig:exploration-modes}
\end{figure}

All three regimes share the same convex backbone and differ only in 
$(S,T,b)$.  In practice we solve the normal equations via Cholesky 
factorization (moderate~$D$) or conjugate gradients (large~$D$), 
increase $\lambda$ or taper $W_S,W_T$ if conditioning deteriorates, 
and relinearize after each step to maintain first-order accuracy.

%%%%%%%%%%%%%%%%%%%%%%%%%%%%%%%%%%

\section{Validation and Experimental Analysis}

We validate MUSIC across three datasets of increasing complexity.
\emph{Gaussian mixtures} (Section~\ref{subsec:gmm_experiment}) provide
a controlled environment in which ground-truth geometry is known,
allowing us to verify inversion accuracy, stability, and local
correctness of the Jacobian model.
\emph{MNIST} (Section~\ref{subsec:mnist}) tests whether the geometric
machinery produces semantically meaningful interpolations on a real
discrete manifold.
\emph{Labeled Faces in the Wild} (Section~\ref{subsec:faces}) probes
robustness on high-dimensional, entangled visual data.

\subsection{Continuity and Topology--Aware Metrics}
\label{subsec:metrics}

To characterize the dynamical behavior of MUSIC trajectories across different datasets, we employ a set of continuity and topology--aware metrics.  
These quantities capture complementary aspects of the motion induced by the control updates, ranging from fine--grained local smoothness to large--scale geometric coherence.  
Because Self--Organizing Maps (SOMs) define a piecewise--linear partition of the input space, with Voronoi regions governed by individual prototypes, trajectory analysis must account for both continuous evolution within a region and discrete reorientations at region boundaries.  
The following metrics provide a comprehensive description of these phenomena and will be used throughout all experimental setups.

\paragraph{Step--direction continuity.}
A first measure of local smoothness is the cosine similarity between consecutive
update directions,
\[
C_t = 
\frac{\langle \Delta z_t,\, \Delta z_{t+1} \rangle}
{\|\Delta z_t\|\, \|\Delta z_{t+1}\|},
\qquad
\Delta z_t = z_{t+1} - z_t .
\]
Values $C_t \simeq 1$ indicate nearly collinear updates, whereas smaller or negative values reveal local directional reorientations.

\paragraph{Topology--aware continuity.}
Because a SOM induces a piecewise--linear tiling of the input space
\citep{kohonen2001basic,yin2008self},
longer trajectories may cross the boundaries between Voronoi regions associated with different prototypes.
At each crossing, the Jacobian structure changes and the optimal update direction is reoriented.
To separate smooth evolution within a region from abrupt changes caused by boundary crossings, we evaluate all continuity metrics in a \emph{topology--aware} manner conditioned on the evolution of the best--matching unit (BMU).

\subparagraph{Transition rate.}
The \emph{transition rate} quantifies how frequently the BMU changes along the path:
\[
r_{\mathrm{trans}}
=
\frac{1}{T}
\sum_{t=0}^{T-1}
\mathbf{1}\!\left[\mathrm{BMU}(z_{t+1}) \neq \mathrm{BMU}(z_t)\right].
\]
Low values correspond to motion within a single basin of attraction; higher values reflect lattice traversal and frequent region crossings.

\subparagraph{Dwell statistics.}
Between two transitions, the trajectory remains associated with the same prototype.
The number of consecutive steps spent within a region defines a \emph{dwell length} $L_k$.
The median and interquartile range (IQR) of the set $\{L_k\}$ quantify the stability of residence within topological cells:
longer dwell times correspond to stable local exploration, while short dwell times indicate rapid hopping between prototypes.

\subparagraph{Curvature.}
The angular deviation between successive directions,
$\theta_t = \cos^{-1}(C_t)$,
provides a notion of local curvature.
We normalize this by the step length to obtain the curvature per unit step,
\[
\kappa_t = \frac{\theta_t}{\|\Delta z_t\|}.
\]
We report median curvature separately for within--cell motion
($\kappa_{\mathrm{within}}$)
and across--cell transitions
($\kappa_{\mathrm{trans}}$).
Large $\kappa_{\mathrm{trans}}$ indicates the sharp reorientations produced at Voronoi boundaries.

\subparagraph{Geodesic efficiency.}
To assess global coherence, we compare the total path length with the direct displacement between start and end points:
\[
E_g
=
\frac{\|z_T - z_0\|}
{\sum_{t=0}^{T-1} \|\Delta z_t\|}.
\]
A value $E_g = 1$ corresponds to a perfectly straight trajectory, while smaller values indicate folded or meandering paths.  
Geodesic efficiency thus complements local metrics by quantifying long--range smoothness.

\subparagraph{Segmented continuity.}
Finally, for each dwell segment (i.e., the interval between two BMU transitions), we compute the average step--direction continuity $\bar C_k$.
The distribution of $\{\bar C_k\}$ captures the internal coherence of motion within stable regions.
We report its median and interquartile range to characterize the variability of local behavior along the trajectory.

\medskip
All continuity--based metrics (step continuity, curvature, geodesic efficiency, and segmented continuity) are unitless and invariant under uniform rescaling of the step sizes.

\subsection{Synthetic validation on Gaussian mixture models}
\label{subsec:gmm_experiment}

We first assess the proposed inversion and interpolation mechanisms in a
controlled setting where both the data distribution and the prototype geometry
are explicitly known.
We consider a three--component Gaussian Mixture Model (GMM) in $\mathbb{R}^D$
(here $D=10$), whose component means forming a triangle in the first two
coordinates, while the remaining dimensions contain small isotropic noise.
From 25{,}000 samples we train a rectangular $20\times 20$ SOM; an independent
test set of 8{,}000 points is standardized using the same statistics.
All experiments in this subsection operate directly in the standardized space.

Given a test point $z$ and its best-matching prototype $r$, the inversion
procedure constructs the anchored linear system $Bz=c$ from squared-distance
activations (Section~\ref{subsec:theory_inversion}), where each row of $B$
encodes an affine difference between prototype $r$ and another prototype.
Using a subset of $N$ prototypes yields an $N\times D$ system.
For each test point and for $N \in \{1,\dots,N_{\max}\}$, we form the
least-squares estimate $\hat z = B^{+}c$ and record the reconstruction error
$\|z - \hat z\|_2$.
Theory predicts that inversion is unstable when $N<D$ (under-determined system)
and becomes exact as soon as $B$ attains full column rank, i.e., once the
selected prototypes span a $D$-dimensional affine subspace.

\begin{figure}[t]
    \centering
    \includegraphics[width=0.72\linewidth]{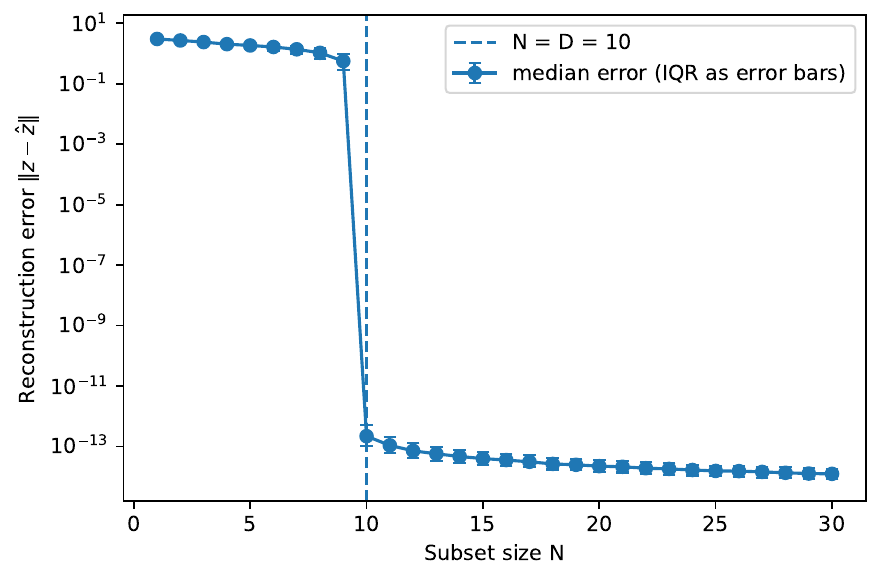}
    \caption{\textbf{Inversion accuracy on a Gaussian mixture.}
    Median reconstruction error $\|z - \hat z\|_2$ as a function of the subset
    size $N$ used to build the anchored system $Bz=c$ from squared-distance
    activations.
    A $20\times 20$ SOM is trained on a $D=10$ GMM; errors are averaged over
    1{,}500 test points, with error bars denoting the interquartile range.
    The dashed vertical line marks the intrinsic dimension $D$.
    In line with the theoretical analysis, inversion is unstable for $N<D$ but
    collapses to machine precision at $N=D$, and remains at the numerical floor
    for all overdetermined systems $N>D$, confirming that squared activations
    carry enough geometric information to recover the input uniquely.}
    \label{fig:gmm_inversion_vs_N}
\end{figure}

Figure~\ref{fig:gmm_inversion_vs_N} shows the median error across 1{,}500 test
points as $N$ increases.
The error remains large and slowly decreasing for $N<D$, then drops
abruptly to numerical zero at $N=D$ and stays there for all $N>D$.
This provides a direct empirical confirmation that the distance-based linear
system is essentially lossless once a full-rank set of prototypes is used.
This outcome is expected: since squared-distance activations are
quadratic in~$z$, the anchored system of activation differences
is exactly linear, and the Jacobian model introduces no
approximation error (Section~\ref{sec:PerturbingSOM}).

Beyond static inversion, the same GMM setting also allows us to study the
\emph{dynamics} induced by MUSIC in a structured, multi-modal manifold.
On this triangular mixture, the trained SOM organizes its prototypes into three
regions aligned with the mixture components.
We then examine two complementary trajectory regimes:

\paragraph{(a) Informed convergence toward a single prototype.}
In the first regime, all trajectories are attracted toward a fixed SOM
prototype chosen near the mean of one GMM component.
Initial states are sampled from all three clusters, so starting points are
topologically diverse.
The MUSIC update rule generates sequences that move toward the selected
prototype (with optional small Gaussian noise).
Despite crossing multiple Voronoi boundaries, trajectories remain locally
smooth with low within--cell curvature and converge to the same attractor,
illustrating a globally coherent flow field capable of targeting isolated
semantic anchors.

\paragraph{(b) Local exploration of a cluster--specific neighbourhood.}
In the second regime, the target at each step is chosen at random among the
prototypes belonging to a selected cluster.
Trajectories again start from all three mixture components: they are first
pulled into the corresponding prototype region and then wandered within it,
showing short dwell lengths, frequent BMU transitions, and moderate curvature.
This behavior reflects a multi--scale organization: MUSIC guides points into the
appropriate semantic basin at a global scale, while locally it induces
piecewise--linear exploration constrained by the SOM topology.

\begin{figure}[t]
    \centering
    \begin{subfigure}[b]{0.48\textwidth}
        \centering
        \includegraphics[width=\linewidth]{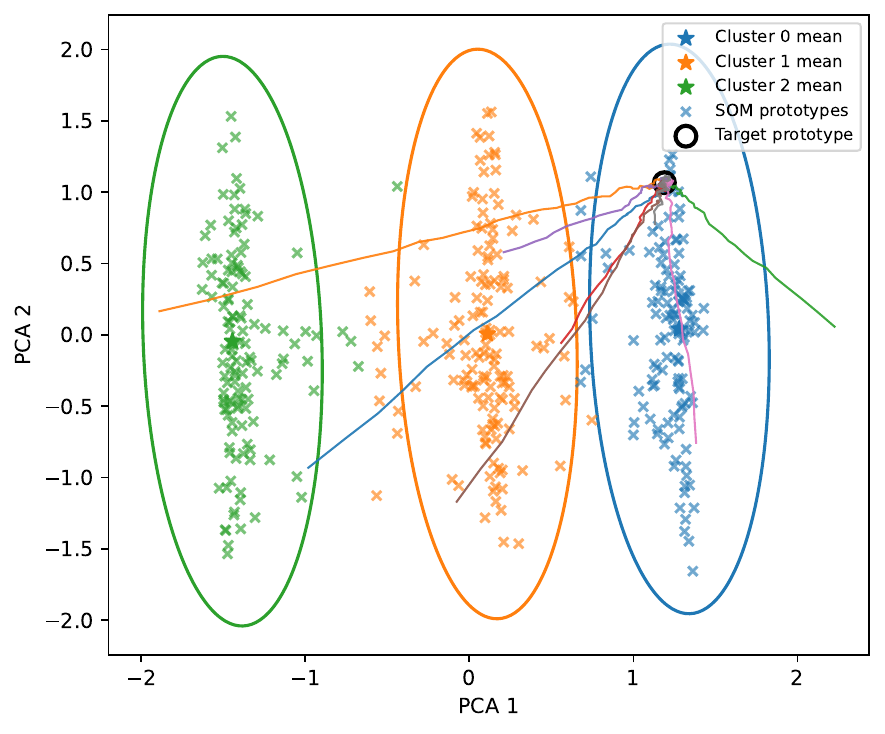}
        \caption{Informed convergence}
        \label{fig:gmm_informed}
    \end{subfigure}\hfill
    \begin{subfigure}[b]{0.48\textwidth}
        \centering
        \includegraphics[width=\linewidth]{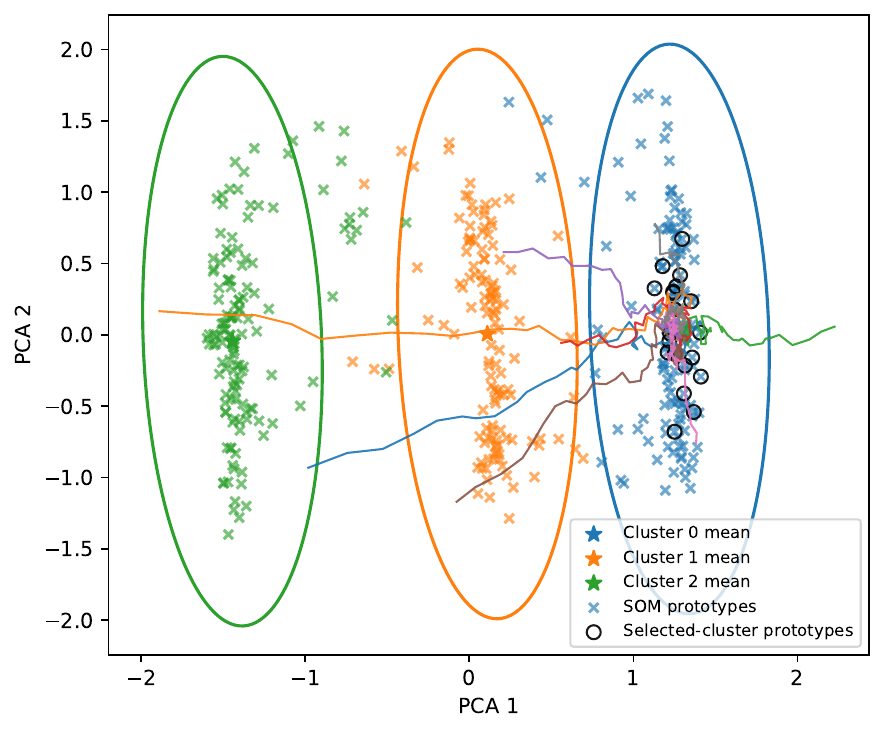}
        \caption{Cluster exploration}
        \label{fig:gmm_exploration}
    \end{subfigure}
    \caption{\textbf{Trajectory behavior on a three--cluster GMM manifold.}
    A SOM is trained on a triangular GMM, yielding three prototype regions
    (colours indicate assignment to the nearest mixture component).
    \textbf{(a)}~MUSIC trajectories initialized from all clusters converge toward
    a single target prototype (circled), following smooth paths with low
    curvature even across Voronoi boundaries.
    \textbf{(b)}~When the target at each step is sampled randomly from the prototypes of one cluster, trajectories are first attracted into that mode
    and then follow short, piecewise--linear exploratory segments within it.
    Ellipses show the $2\sigma$ covariance contours of the three GMM components.}
    \label{fig:gmm_trajectory_experiments}
\end{figure}

To complement the qualitative trajectory plots, we report in
Table~\ref{tab:gmm_trajectory_metrics} a set of global scalar metrics that
summarize the behaviour of MUSIC dynamics on the GMM manifold.  We focus on the topology-aware measures defined in Section~\ref{subsec:metrics}: transition rate, dwell statistics, step--direction continuity, curvature, geodesic efficiency, and segmented continuity.

\begin{table*}[t]
\centering
\begin{subtable}[t]{0.48\textwidth}
    \centering
    \begin{tabular}{lc}
    \toprule
    Metric & Median $\pm$ IQR \\
    \midrule
    Transition rate $r_{\mathrm{trans}}$        & $0.042 \pm 0.021$ \\
    Dwell length (steps)                        & $6.0 \pm 10.25$ \\
    Dwell IQR                                   & $17.38 \pm 12.50$ \\
    Step--direction continuity $C$               & $0.291 \pm 0.093$ \\
    Curvature $\kappa$                          & $34.86 \pm 7.12$ \\
    Geodesic efficiency $E_g$                   & $0.757 \pm 0.089$ \\
    Segmented continuity $\bar C_k$             & $0.944 \pm 0.362$ \\
    Segmented continuity IQR                    & $0.364 \pm 0.188$ \\
    \bottomrule
    \end{tabular}
    \caption{Informed convergence}
    \label{tab:gmm_metrics_informed}
\end{subtable}\hfill
\begin{subtable}[t]{0.48\textwidth}
    \centering
    \begin{tabular}{lc}
    \toprule
    Metric & Median $\pm$ IQR \\
    \midrule
    Transition rate $r_{\mathrm{trans}}$        & $0.383 \pm 0.100$ \\
    Dwell length (steps)                        & $1.5 \pm 1.13$ \\
    Dwell IQR                                   & $1.75 \pm 1.44$ \\
    Step--direction continuity $C$               & $0.079 \pm 0.274$ \\
    Curvature $\kappa$                          & $9.42 \pm 0.93$ \\
    Geodesic efficiency $E_g$                   & $0.291 \pm 0.059$ \\
    Segmented continuity $\bar C_k$             & $-0.007 \pm 0.328$ \\
    Segmented continuity IQR                    & $0.475 \pm 0.165$ \\
    \bottomrule
    \end{tabular}
    \caption{Cluster exploration}
    \label{tab:gmm_metrics_exploration}
\end{subtable}
\caption{\textbf{Quantitative analysis of MUSIC trajectories on the GMM manifold.}
Each entry reports the median $\pm$ IQR across trajectories.
\textbf{(Left)}~Informed convergence: trajectories exhibit low transition rate,
long dwell times, higher global step--direction continuity, and high geodesic
efficiency, indicating globally directed motion toward a single prototype.
\textbf{(Right)}~Cluster exploration: trajectories undergo frequent BMU
transitions, short dwell times, lower continuity, and reduced geodesic
efficiency, reflecting locally wandering, topology--constrained exploration
within the selected mixture component.}
\label{tab:gmm_trajectory_metrics}
\end{table*}

The metrics reveal a clear contrast between the two dynamical regimes.  
In the informed--convergence setting, the transition rate is low and dwell
times are long, reflecting coherent motion within stable Voronoi regions.
Step--direction continuity remains high and geodesic efficiency is close to~1,
indicating globally directed flow toward the designated prototype.
In the cluster--exploration regime, transitions are frequent and dwell times
shrink to one or two steps.
Continuity decreases, curvature becomes more variable, and geodesic efficiency
drops substantially, consistent with locally wandering, topology--constrained
motion within the selected mixture component.
Together, these experiments show how MUSIC behaves on a simple multimodal
manifold: through \emph{directed} convergence to a single prototype and
\emph{exploratory} motion confined to a cluster--specific region.

\subsection{Experiments on MNIST}
\label{subsec:mnist}

\begin{figure}[t]
    \centering
    \includegraphics[width=0.55\linewidth]{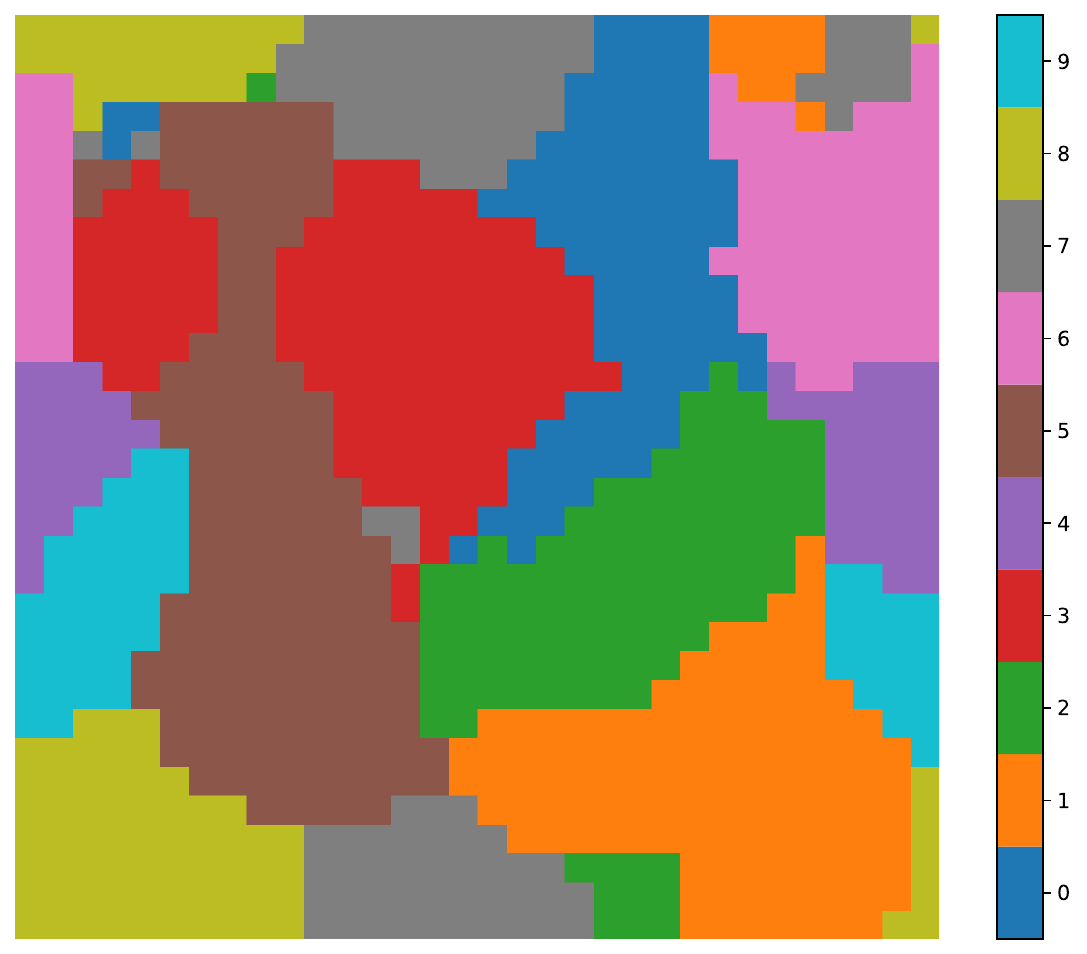}
    \caption{\textbf{Prototype organization on MNIST.}
    A $32\times 32$ toroidal SOM trained on PCA--whitened MNIST forms ten coherent
    prototype regions corresponding to the digit classes.
    Neighbouring prototypes capture smooth morphological variations, while
    class boundaries emerge along contiguous regions of the map.}
    \label{fig:mnist_som_map}
\end{figure}

We train a $32 \times 32$ toroidal SOM on the MNIST handwritten digit dataset
\citep{lecun1998mnist} after full--dimensional PCA whitening.
The toroidal topology avoids boundary artefacts and forces all classes to be
embedded in a common continuous latent surface.
By assigning each prototype the majority label of its BMU matches, we obtain
ten coherent prototype regions corresponding to the ten digit classes
(Fig.~\ref{fig:mnist_som_map}).
Neighbouring prototypes within each region represent smooth morphological
variations (stroke thickness, orientation, style), while transitions between
classes occur along contiguous boundaries where prototype basins meet.
This topological structure provides the latent geometry on which MUSIC
trajectories operate.

\subsubsection{Inversion behaviour on MNIST}

Before analysing MUSIC trajectories, we verify that the inversion mechanism
derived in Section~\ref{subsec:theory_inversion} remains numerically stable on
real data.
For each MNIST test image, we compute squared--distance activations from the SOM
prototypes, form the anchored system $Bz=c$, and recover $\hat z = B^{+}c$.
Because the map contains many more than $D$ prototypes spanning the
whitened PCA space in general position, the systems are highly
overdetermined, and reconstruction errors $\|z-\hat z\|_2$ are consistently
near machine precision.
This guarantees that intermediate MUSIC states can be faithfully decoded
back into pixel space.

\subsubsection{Directed class transition}

\begin{figure}[t]
    \centering
    \includegraphics[width=1.0\linewidth]{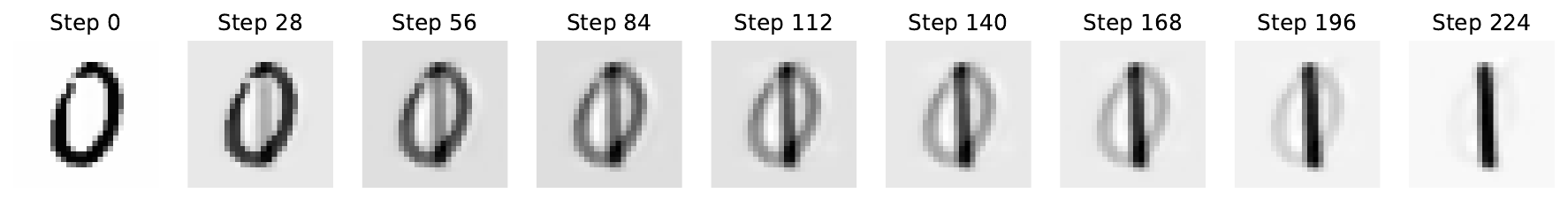}
    \caption{\textbf{MUSIC trajectory from ``0'' to ``1'' on MNIST (Tikhonov variant).}
Starting from a test image of the digit ``0'' (left), MUSIC is run in informed mode toward a target prototype representing the digit ``1'' (right) using the
Tikhonov-regularized update rule.  
The intermediate reconstructions exhibit a smooth and interpretable evolution: the circular stroke of the ``0'' gradually fades while a vertical segment
progressively appears and sharpens.
Around the middle of the path, both structures are partially visible, forming a hybrid configuration that reflects the proximity of the two class regions on
the SOM.
These intermediate states result from manifold-aware latent updates rather than
pixel-space interpolation, and the Tikhonov formulation yields a particularly
stable, smooth transition.}

    \label{fig:mnist_MUSIC_transition_0_1}
\end{figure}

We analyse a cross--class MUSIC interpolation from a digit ``0'' toward a
prototype representing ``1''
(Fig.~\ref{fig:mnist_MUSIC_transition_0_1}).
Initially, the updates deform the ``0'' while preserving its circular outline.
As the trajectory approaches the boundary between the ``0'' and ``1'' prototype
regions, a hybrid configuration emerges: the loop becomes fainter while a
vertical stroke begins to appear.
After crossing the boundary, the trajectory aligns with the ``1'' region and
converges smoothly toward the target prototype.

\begin{figure}[t]
    \centering
    \includegraphics[width=0.75\linewidth]{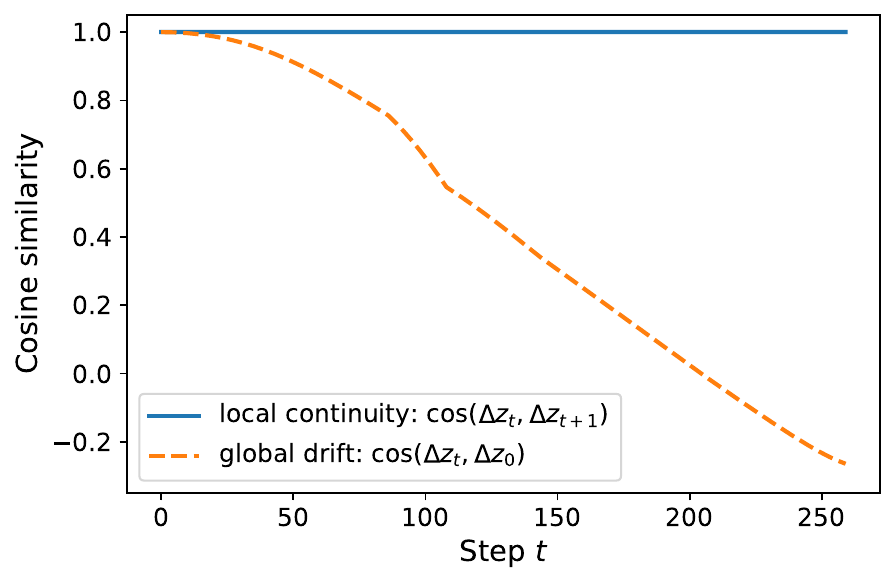}
    \caption{\textbf{Step--direction continuity on the MNIST ``0''$\to$``1''
    transition.}
    We report the cosine similarity between successive update directions,
    $\cos(\Delta z_t, \Delta z_{t+1})$, and between each update and the first
    one, $\cos(\Delta z_t, \Delta z_{0})$, along the informed trajectory.
    The high local similarity (solid line) shows that consecutive steps remain
    nearly collinear, confirming the smoothness of MUSIC updates in the latent
    space.
    The gradual decay of global similarity (dashed line) reflects the expected
    reorientation of the trajectory as it crosses multiple Voronoi regions on
    the SOM while moving from the ``0'' manifold toward the ``1'' prototype.
    Together, these curves quantify the locally coherent yet globally bending
    nature of the MUSIC flow on a real data manifold.}
    \label{fig:mnist_cosine_continuity}
\end{figure}

To quantify the smoothness of the latent trajectory beyond visual inspection (Fig.~\ref{fig:mnist_cosine_continuity}), we compute cosine similarities between successive update directions,
$\cos(\Delta z_t, \Delta z_{t+1})$, and between each update and the initial
direction, $\cos(\Delta z_t, \Delta z_0)$.
The first remains consistently close to~1 throughout, indicating that
consecutive steps are almost perfectly aligned.
The second decreases steadily, revealing a gradual reorientation as the path
crosses several Voronoi boundaries.
This illustrates the expected two-scale structure of MUSIC: locally coherent,
low-curvature motion within Voronoi regions, together with a global bending
as the trajectory moves from the ``0'' region toward the ``1'' prototype.

% -----------------------------------------------------------------------
% NEW MNIST EXPERIMENTAL SUBSUBSECTIONS
% Insert after the cosine continuity figure (\ref{fig:mnist_cosine_continuity})
% in Section 5.4 (Experiments on MNIST)
% -----------------------------------------------------------------------

\subsubsection{Classifier confidence along trajectories}
\label{subsubsec:classifier_confidence}

To assess whether intermediate MUSIC states remain semantically valid,
we train an MLP classifier in the SOM's PCA-whitened latent space
(validation accuracy $98.0\%$) and evaluate the predicted class
probabilities at every step along the $0\!\to\!1$ trajectory of
Fig.~\ref{fig:mnist_MUSIC_transition_0_1}.
We compare MUSIC with a linear interpolation baseline that moves
along the straight line from the source image to the target prototype
in the same latent space.

\begin{figure}[t]
    \centering
    \includegraphics[width=0.85\linewidth]{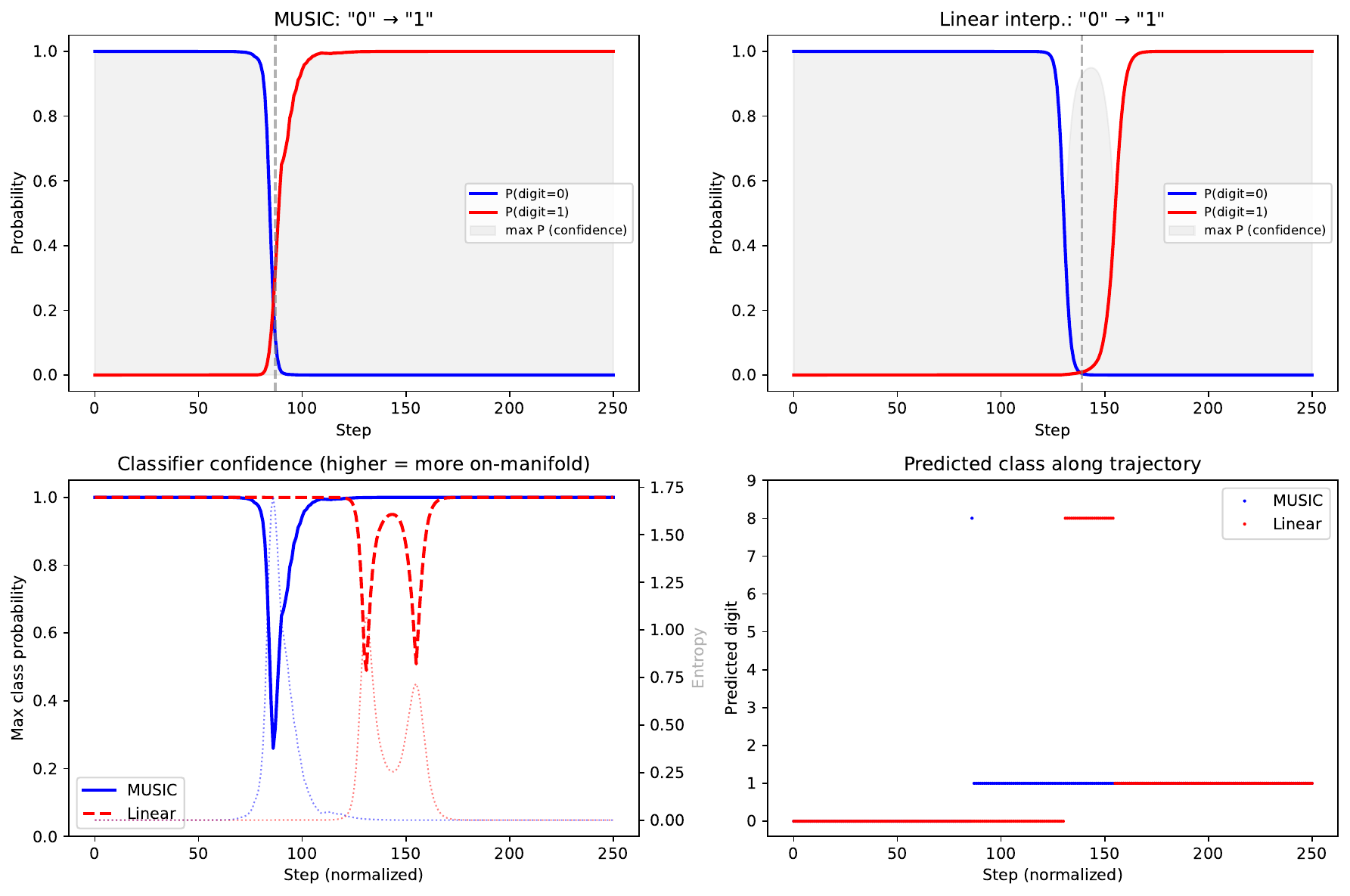}
    \caption{\textbf{Classifier confidence along the $0\!\to\!1$ trajectory.}
    An MLP classifier (validation accuracy $98.0\%$) evaluates the predicted
    class probabilities at each step.
    MUSIC (left) transitions from ``0'' to ``1'' at step~87 via a single
    boundary step through ``8'', while linear interpolation (right)
    lingers in the ``8'' region for 24 consecutive steps before reaching
    ``1'' at step~155.}
    \label{fig:classifier_confidence}
\end{figure}

Figure~\ref{fig:classifier_confidence} shows the resulting confidence
profiles.
Both methods maintain high mean confidence ($0.974$ for MUSIC, $0.973$
for linear), but differ sharply in their transition dynamics.
MUSIC crosses from the source to the target class at step~87
($34.8\%$ of the trajectory), whereas linear interpolation crosses
only at step~139 ($55.6\%$).
More revealing is the sequence of predicted labels.
MUSIC follows a clean $0\!\to\!0\!\to\!1\!\to\!1$ progression, passing
through the ``8'' class for a single boundary step before committing
to the target.
Linear interpolation, by contrast, enters the ``8'' region at step~131
and remains there for 24 consecutive steps before finally reaching ``1''
at step~155.
This prolonged detour through a semantically unrelated class reflects
the straight-line path cutting across the SOM lattice without regard
for prototype topology, while MUSIC follows the Voronoi structure
and transitions directly between neighbouring class basins.

\subsubsection{Comparison with VAE interpolation}
\label{subsubsec:vae_comparison}

We compare MUSIC trajectories with latent-space interpolation in a
Variational Autoencoder (VAE) trained on the same MNIST data
(32-dimensional latent space, 30 epochs).
For each trajectory from the ``0'' source to the ``1'' target, we
measure two complementary metrics: the mean $k$-nearest-neighbour
distance to the training set (manifold proximity) and the mean
Laplacian sharpness of the reconstructed images.

The VAE stays marginally closer to the data manifold (mean distance
$3.93$ vs.\ $4.08$ for MUSIC and $4.10$ for linear interpolation),
owing to its learned decoder prior which pulls outputs toward the
training distribution.
However, MUSIC produces the sharpest intermediate images
(sharpness $0.051$ vs.\ $0.038$ for VAE and $0.047$ for linear),
reflecting the fact that MUSIC reconstructions are anchored to
specific prototypes rather than smoothed by a decoder bottleneck.
This tradeoff---slightly farther from the data centroid but
structurally sharper---is consistent with MUSIC operating on the
discrete SOM geometry rather than on a smooth probabilistic manifold.

\subsubsection{Systematic evaluation across digit pairs}
\label{subsubsec:systematic_eval}

To confirm that the preceding observations generalise beyond a single
pair, we evaluate MUSIC and linear interpolation across ten diverse
digit pairs
($0\!\to\!1$, $1\!\to\!7$, $2\!\to\!3$, $3\!\to\!8$, $4\!\to\!9$,
$5\!\to\!6$, $6\!\to\!0$, $7\!\to\!2$, $8\!\to\!5$, $9\!\to\!4$),
with three independent starting images per pair, yielding 60
trajectories in total.
Table~\ref{tab:systematic_mnist} summarises the results.

\begin{table}[t]
\centering
\caption{\textbf{Systematic comparison of MUSIC vs.\ linear
interpolation on MNIST} (10 digit pairs $\times$ 3 samples = 30
trajectories per method).
Each cell reports median $\pm$ IQR.
Bold indicates the statistically better method (Wilcoxon signed-rank
test, $p<0.05$); ``ns'' denotes no significant difference.}
\label{tab:systematic_mnist}
\small
\begin{tabular}{lccl}
\toprule
Metric & MUSIC & Linear & Sig.\ \\
\midrule
Mean confidence
  & $0.978 \pm 0.053$
  & $0.987 \pm 0.012$
  & ns ($p\!=\!0.70$) \\[2pt]
Min confidence
  & $\mathbf{0.728 \pm 0.224}$
  & $0.615 \pm 0.209$
  & $8/10$ pairs \\[2pt]
Mean sharpness
  & $\mathbf{0.146 \pm 0.035}$
  & $0.117 \pm 0.020$
  & $p\!=\!0.002$ \\[2pt]
Manifold distance
  & $7.11 \pm 0.86$
  & $\mathbf{5.94 \pm 0.52}$
  & $p\!=\!0.002$ \\[2pt]
Geodesic efficiency
  & $0.738 \pm 0.094$
  & $1.000 \pm 0.000$
  & --- \\[2pt]
Step continuity
  & $0.942 \pm 0.116$
  & $1.000 \pm 0.000$
  & --- \\
\bottomrule
\end{tabular}
\end{table}

Mean classifier confidence is comparable between the two methods
($p\!=\!0.70$), confirming that MUSIC trajectories stay within
recognisable digit regions to the same degree as straight-line paths.
MUSIC achieves significantly higher minimum confidence, winning
on 8 out of 10 pairs: the worst-case intermediate frame is more
recognisable with MUSIC than with linear interpolation.
Image sharpness favours MUSIC on all 10 pairs ($p\!=\!0.002$),
consistent with the prototype-anchored nature of the updates.

The tradeoff is path length: MUSIC's geodesic efficiency averages
$0.74$, reflecting the piecewise-linear routing through Voronoi
regions.
Linear interpolation achieves $E_g\!=\!1$ by definition, but at the
cost of the semantic detours documented above.
Mean manifold distance is accordingly higher for MUSIC ($7.11$ vs.\
$5.94$, $p\!=\!0.002$), because the curved path visits a broader
portion of the latent space rather than cutting through low-density
regions.

\subsubsection{Hyperparameter sensitivity}
\label{subsubsec:hyperparam}

We sweep the main MUSIC parameters on the $0\!\to\!1$ trajectory
(150~steps) to characterise the stability of the method.

\paragraph{Preservation weight $\gamma$.}
Varying $\gamma$ over $[0.5,\,0.99]$ while holding
$\lambda\!=\!2\!\times\!10^{-4}$ and $\eta\!=\!0.08$, mean classifier
confidence remains between $0.950$ and $0.958$, confirming that the
method is robust to this parameter.
Step-direction continuity decreases moderately from $0.70$ to $0.64$
and geodesic efficiency increases from $0.60$ to $0.75$ as $\gamma$
grows, reflecting the expected shift from exploration toward
target-directed motion.

\paragraph{Tikhonov parameter $\lambda$.}
The regularisation strength $\lambda$ exhibits textbook behaviour.
At very low regularisation ($\lambda\!=\!10^{-6}$), confidence reaches
$0.9999$ but continuity drops to $-0.12$, indicating erratic,
high-frequency steps that remain on-manifold but lack directional
coherence.
As $\lambda$ increases from $10^{-5}$ to $10^{-3}$, continuity rises
from $0.43$ to $0.67$ while confidence stabilises around $0.95$,
yielding the recommended operating range.
At $\lambda\!=\!10^{-1}$, continuity reaches $0.999$ and the flow
becomes ultra-smooth but nearly stationary (convergence ratio $0.03$).
This validates the L-curve heuristic suggested in
Section~\ref{sec:MUSIC}.

\paragraph{Target gain $\eta$.}
In the informed (single-target) mode, $\eta$ has no measurable effect:
all metrics remain identical across $[0.01,\,0.20]$.
This is expected from the theory: with a single target row in $B_T$,
$\eta$ scales only the magnitude of $\Delta z$, which is subsequently
clipped by the step-size parameter $\rho$.
The direction of the update is therefore invariant to $\eta$; this parameter becomes relevant only in cluster mode, where it
balances the contributions of multiple target prototypes.

\subsection{Experiments on Faces in the Wild}
\label{subsec:faces}

We use a ``Faces in the Wild'' dataset \citep{huang2008labeled}, encode each
image into a 512--dimensional latent vector via a pretrained autoencoder, and apply PCA whitening.
A $32\times32$ SOM trained on these latent codes organizes faces according to broad semantic attributes---gender, skin tone, hairstyle, facial structure, and
illumination---yielding a rich topological landscape suitable for controlled
semantic explorations.

\paragraph{Free evolution.}
To illustrate the semantic benefits of geometry-aware inversion, we perform a free-evolution experiment starting from the same reference face. At each step, we randomly select a SOM prototype and impose a random perturbation of its squared distance. In the \emph{MUSIC} condition, the update direction is obtained via our manifold-consistent inversion (preserving a small set of distances while driving the selected one), and the latent step is normalized to a fixed magnitude. In the \emph{no-constraints} baseline, the latent update is instead taken along the purely radial direction of that prototype, without any geometric preservation. 

\begin{figure}[t]
    \centering
    \includegraphics[width=1.0\linewidth]{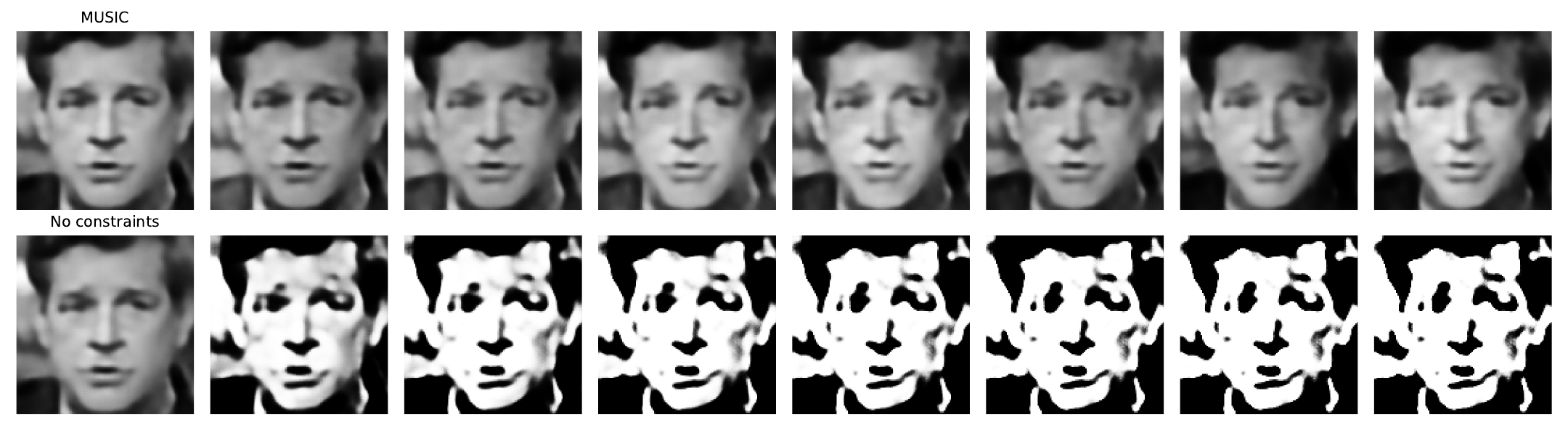}
    \caption{\textbf{Free evolution in latent space under geometry-aware control.} 
Starting from the same input face (left), we repeatedly apply random perturbations to the squared distance of a randomly selected SOM prototype. In the \emph{MUSIC} row (top), each perturbation is processed through manifold-aware inversion: the update direction is chosen to satisfy the perturbed distance while preserving a small subset of nearby prototype distances, and the resulting step is normalized to a fixed magnitude. In the \emph{no-constraints} row (bottom), the latent update is taken along the purely radial direction of the selected prototype, with the same step length but without any geometric preservation. 
Despite identical latent step sizes, the unconstrained trajectory rapidly drifts off the face manifold and produces distorted images, whereas MUSIC yields smooth, semantically coherent transformations that preserve the overall facial structure.}

    \label{fig:free_evolution_faces}
\end{figure}

Although both trajectories have identical step lengths in latent space, their qualitative behaviour differs drastically. The unconstrained trajectory rapidly drifts into implausible regions, yielding distorted faces. In contrast, the MUSIC trajectory evolves smoothly and remains on the face manifold, exhibiting coherent semantic transformations without collapsing into artifacts (Fig.~\ref{fig:free_evolution_faces}). For a more detailed description of this experiment, refer to Supplementary Section~\ref{supp:S6}.

\paragraph{Informed exploration toward semantic clusters.}
Beyond local transformations, MUSIC exhibits structured global behavior: when iterated with consistent prototype selections, the latent trajectory converges toward the corresponding cluster.
We run MUSIC in informed mode, where at each step the update is directed toward a single prototype selected from a user--chosen target cluster.
This one--prototype--at--a--time guidance produces smooth, decoder--faithful trajectories that remain close to the learned manifold.
A key property is identity preservation: because each step is constrained by the local SOM geometry, core identity cues remain stable until the trajectory approaches the target region, while attributes such as skin tone or facial softness evolve gradually.

\begin{figure}[t]
    \centering
    \begin{subfigure}[b]{0.95\linewidth}
        \centering
        \includegraphics[width=\linewidth]{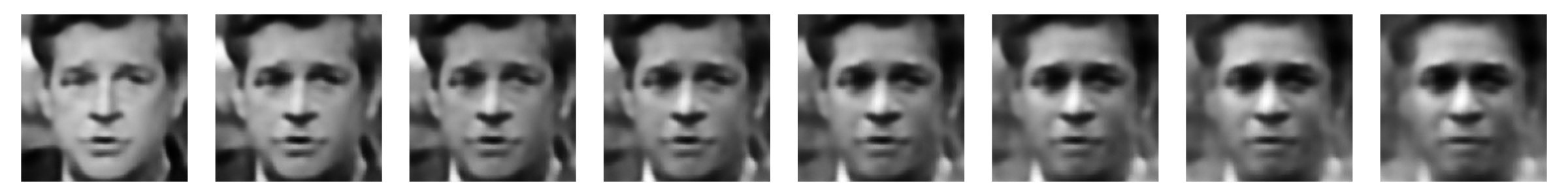}
        \caption{Toward ``black faces'' cluster}
    \end{subfigure}
    \vfill
    \begin{subfigure}[b]{0.95\linewidth}
        \centering
        \includegraphics[width=\linewidth]{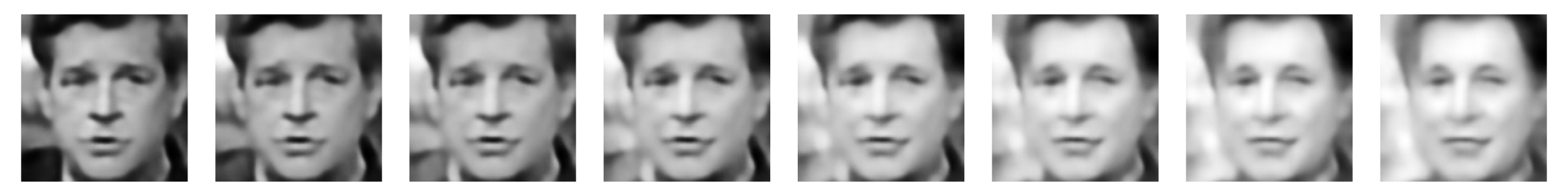}
        \caption{Toward ``female faces'' cluster}
    \end{subfigure}
    \vfill
    \begin{subfigure}[b]{0.95\linewidth}
        \centering
        \includegraphics[width=\linewidth]{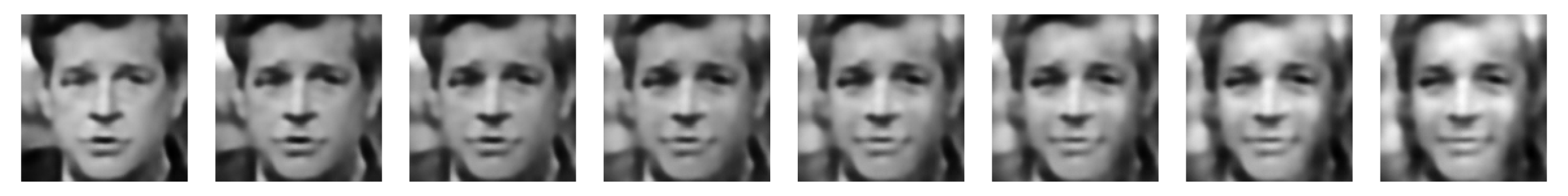}
        \caption{Toward 5-prototype custom set}
    \end{subfigure}
    \caption{\textbf{MUSIC informed explorations on the facial manifold.}
    Starting from the same validation image (leftmost column), MUSIC moves along
    smooth, semantically meaningful directions induced by SOM topology.
    Each trajectory is guided by repeatedly selecting a single prototype from
    the target set. The input face is taken from the held-out validation set and was not used during SOM training.}
    \label{fig:faces_experiments}
\end{figure}

We illustrate three explorations, each probing a distinct semantic
direction on the SOM (Fig.~\ref{fig:faces_experiments}):

\paragraph{(a) Transition toward a demographic region.}
Starting from a validation image, we guide the trajectory toward a cluster
containing predominantly darker--skinned faces.
The evolution shows a smooth darkening of skin tone and gradual adjustments in
local shading and facial texture, while identity remains stable deep into
the trajectory.

\paragraph{(b) Transition toward a gender--specific region.}
Using the same initial face, we target a cluster of mainly female prototypes.
The resulting path progressively introduces features associated with the
cluster---subtle changes in eyes, eyebrows, and facial softness---without
abrupt distortions.

\paragraph{(c) Convergence toward a small user--defined prototype set.}
To test MUSIC in a minimally constrained setting, we select five
prototypes at random (not forming a labelled cluster) and use them as informed
anchors.
The trajectory remains stable and interpretable, moving toward a mixed
semantic region combining attributes of the selected prototypes.

\begin{figure}[t]
    \centering
    \includegraphics[width=0.75\linewidth]{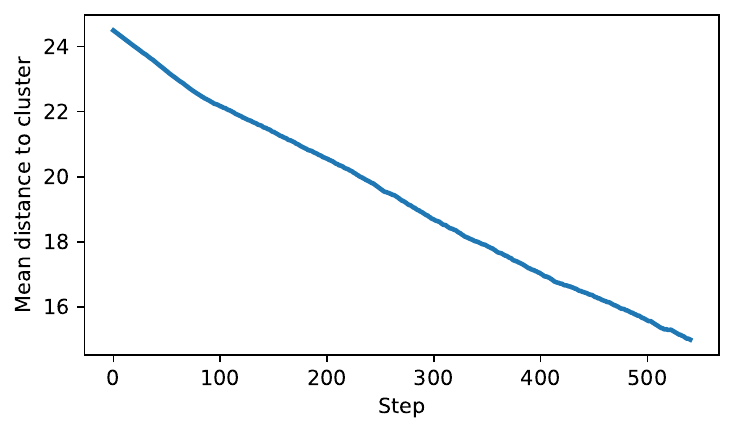}
    \caption{\textbf{Convergence toward the target prototype cluster.}
    Mean Euclidean distance between $z_t$ and the ``female'' cluster prototypes
    along the trajectory of Fig.~\ref{fig:faces_experiments}(b).
    The monotonic decay confirms directed motion toward the target region.}
    \label{fig:faces_cluster_distance}
\end{figure}

To quantitatively verify that informed trajectories reach their intended
semantic regions, we measure---for the experiment targeting the ``female''
cluster---the mean Euclidean distance between the latent state $z_t$ and all
prototypes in that cluster.
As shown in Fig.~\ref{fig:faces_cluster_distance}, the distance decreases
smoothly and monotonically, confirming that the latent dynamics consistently
move toward the target region.

\paragraph{Comparison with linear interpolation.}
For a single face--to--prototype transition, linear interpolation in the 512-dimensional latent space produces very similar results, indicating that the autoencoder space is already locally Euclidean along this direction.
We therefore consider a stronger baseline: linear interpolation from the initial face to the \emph{mean} of the target cluster (the same female-faces cluster used above).

This comparison isolates two different mechanisms.
Linear interpolation moves straight toward the cluster barycenter, effectively averaging all facial features and producing increasingly blurred, identity-agnostic faces.
Informed MUSIC instead follows a piecewise-smooth trajectory guided by actual prototypes; every intermediate point remains close to a real face embedding, preserving structure and identity.

\begin{figure}[t]
    \centering
    \includegraphics[width=1.0\linewidth]{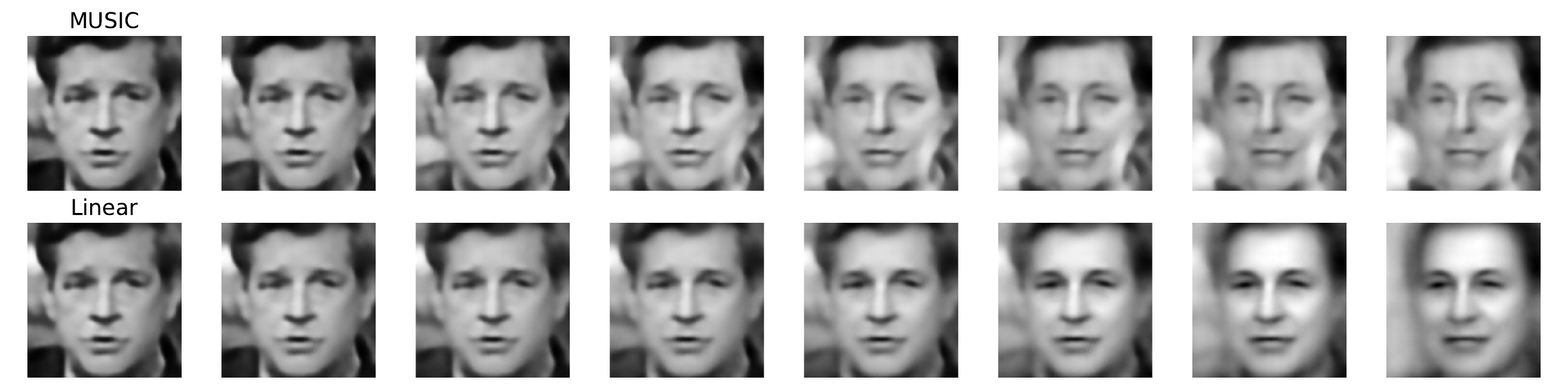}
    \caption{\textbf{MUSIC vs.\ linear interpolation toward a target facial cluster.}
    MUSIC (top row) preserves identity and produces sharp intermediate faces,
    while linear interpolation (bottom row) collapses toward a blurred cluster mean.
    Both trajectories target the female-faces cluster.}
    \label{fig:cluster_transition_MUSIC_vs_linear_images}
\end{figure}

\begin{figure}[t]
    \centering
    \includegraphics[width=1.0\linewidth]{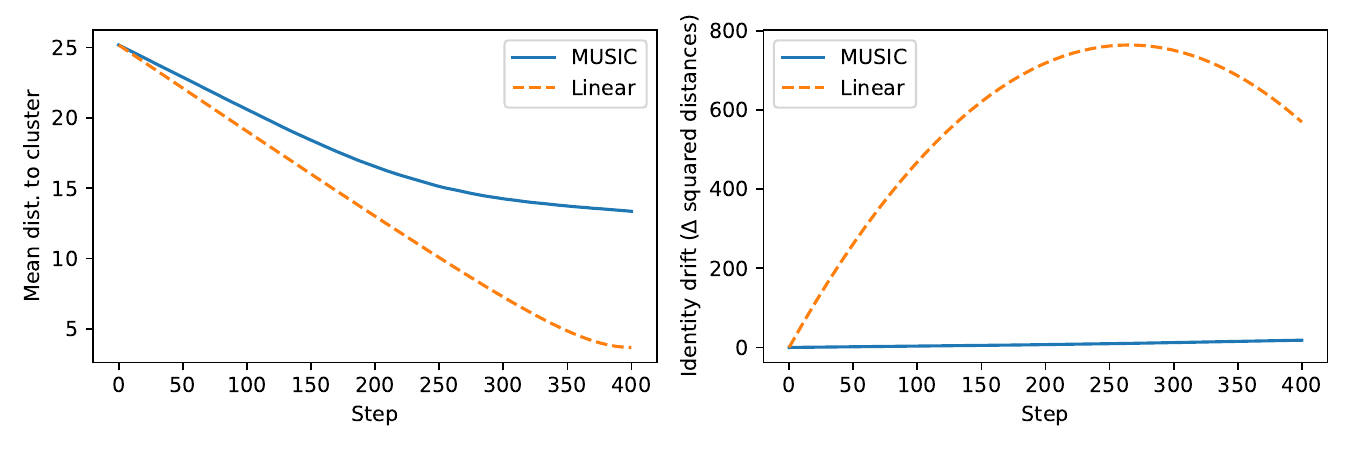}
    \caption{\textbf{Convergence and identity drift during MUSIC vs.\ linear transition.}
    (\emph{Left}) Mean distance to the female-faces cluster.  
    Linear interpolation collapses to the cluster barycenter, while MUSIC approaches the cluster more conservatively.  
    (\emph{Right}) Identity drift relative to initial anchors.  
    MUSIC preserves identity for most of the trajectory, whereas linear interpolation causes a dramatic loss (up to $\sim\!800$) before partially recovering near its endpoint.}
    \label{fig:cluster_transition_MUSIC_vs_linear_metrics}
\end{figure}

Figure~\ref{fig:cluster_transition_MUSIC_vs_linear_images} illustrates these effects: the linear path collapses rapidly into a blurry representation, while MUSIC produces sharper, more realistic intermediate faces with the original identity preserved deep into the sequence.
The accompanying metric plots (Fig.~\ref{fig:cluster_transition_MUSIC_vs_linear_metrics}) confirm this quantitatively.
Linear interpolation reduces the cluster distance aggressively (from ${\sim}25$ to ${\sim}3$) but causes identity drift climbing to ${\sim}800$, whereas MUSIC converges more moderately (to ${\sim}15$) while maintaining near-zero identity drift throughout.

For reproducibility, all experiments on faces were performed using a single fixed configuration of MUSIC hyperparameters ($\gamma = 0.85$, $\lambda = 10^{-4}$, $\rho = 0.1$, $\eta = 0.04$, $\sigma = 0.01$), ensuring that the observed behaviours arise from the method itself rather than from per-sample tuning.

Overall, these experiments demonstrate that MUSIC remains stable and
semantically meaningful on a highly nonlinear manifold of real faces.
Despite the complex entanglement of identity, pose, illumination, and 
demographic attributes, informed MUSIC trajectories perform long--range
semantic transitions while preserving photometric consistency and remaining
within the domain of valid decoder embeddings.

%%%%%%%%%%%%%%%%%%%%%%%%%%%%%%%%%%

\section{Discussion}

A central insight of this work is that the structure of the prototype set plays
a crucial role in the stability and interpretability of the resulting
trajectories.  
While the inversion mechanism requires only $N \ge D{+}1$ affinely independent
vectors, the behaviour of MUSIC depends strongly on how these reference points are organized in space.  
If prototypes are sampled randomly or obtained from unstructured clustering, the resulting Voronoi cells have no coherent semantic geometry: neighbourhoods do
not correspond to gradual attribute changes, boundary transitions occur
abruptly, and intermediate states are difficult to interpret.  
In such settings, MUSIC produces unstable or chaotic motion because the underlying partition lacks topological meaning.

In contrast, a Self--Organizing Map provides a two--dimensional,
topology-preserving lattice in which neighbouring prototypes correspond to
semantically similar inputs.  
This induces smooth Voronoi regions and contiguous boundaries that together form
a piecewise--linear coordinate chart of the data manifold.  
MUSIC leverages precisely this structure: within a single region, the
Tikhonov-regularized flow produces low-curvature motion aligned with local
prototype geometry, while boundary crossings occur in a predictable manner
reflecting the lattice topology.  
The systematic evaluation on MNIST confirms this empirically: MUSIC achieves significantly higher image sharpness than linear interpolation on all ten digit pairs tested ($p\!=\!0.002$; Table~\ref{tab:systematic_mnist}), reflecting the prototype-anchored nature of the updates.
This behaviour cannot be replicated by standard clustering methods such as
$k$--means or $k$NN grouping, which lack neighbourhood continuity and provide no
global topological backbone for guiding semantic transformations.

Viewed from this perspective, MUSIC offers a deterministic,
geometry-driven mechanism for semantic interpolation that complements modern
generative models.  
Whereas VAEs or diffusion models define trajectories through neural
parameterisation or stochastic sampling, MUSIC defines them through prototype geometry and linear algebra alone.  
A direct comparison on MNIST illustrates the tradeoff: a VAE trained on the same data stays marginally closer to the data manifold (mean $k$-NN distance $3.93$ vs.\ $4.08$), but MUSIC produces the sharpest intermediate images (Laplacian sharpness $0.051$ vs.\ $0.038$; Section~\ref{subsubsec:vae_comparison}).
This reflects a fundamental distinction: the VAE decoder smooths outputs toward the training distribution, while MUSIC reconstructions remain anchored to discrete prototypes and preserve structural detail.
Both approaches thus provide interpretability---MUSIC through geometric transparency, generative models through probabilistic structure---and may be profitably combined.

A comparison with linear interpolation in the autoencoder latent space further
clarifies the behaviour of MUSIC.
The classifier confidence experiment (Section~\ref{subsubsec:classifier_confidence}) reveals a striking qualitative difference on the $0\!\to\!1$ trajectory: MUSIC transitions directly between the two class basins, passing through the ``8'' boundary region for a single step, whereas linear interpolation lingers in the ``8'' class for 24 consecutive steps before reaching the target.
This pattern generalises across digit pairs: in the systematic evaluation, MUSIC achieves higher minimum confidence on 8 out of 10 pairs (Table~\ref{tab:systematic_mnist}), indicating that its worst-case intermediate frames remain more recognisable.
When moving toward an entire semantic cluster on faces, the advantage becomes even more pronounced:
straight-line interpolation collapses toward the barycentric mean,
producing a blurred and unrealistic face, whereas MUSIC follows prototype transitions and maintains identity much longer (Fig.~\ref{fig:cluster_transition_MUSIC_vs_linear_metrics}).
The price is a longer path through the latent space (geodesic efficiency $0.74$ vs.\ $1.0$), which is the expected cost of following the SOM topology rather than cutting through low-density regions.

From a computational standpoint, each MUSIC update requires solving a
$D$-dimensional linear system of the form
\[
\big((1-\gamma)A^\top A + \gamma B^\top B + \lambda I\big)\,\Delta z = r,
\]
where $A$ and $B$ contain only a small number of Jacobian rows associated with
preserved and target prototypes.  
Since these matrices are extremely low rank relative to $D$, the computational cost is dominated by operations linear in the latent dimension.  
This makes MUSIC substantially lighter than diffusion models or autoregressive architectures, which require a full neural forward pass at every iteration.
Moreover, the method does not require a generative decoder unless visual inspection of intermediate states is desired.

The hyperparameter analysis (Section~\ref{subsubsec:hyperparam}) confirms that the method is robust in practice.
Mean classifier confidence varies by less than one percentage point across the full range of the preservation weight $\gamma \in [0.5, 0.99]$.
The Tikhonov parameter $\lambda$ exhibits textbook regularisation behaviour, with a well-defined operating regime between $10^{-5}$ and $10^{-3}$ where continuity and convergence are jointly favourable; this validates the L-curve heuristic suggested in Section~\ref{sec:MUSIC}.
The target gain $\eta$ is provably irrelevant in single-target mode (the step-size clipping absorbs the scaling), but becomes active in cluster mode where it balances multiple target contributions.

The method naturally inherits some limitations.  
Exact inversion is guaranteed only when the prototypes span the latent space,
which may fail if the intrinsic dimension exceeds the number of prototypes or if prototypes are poorly distributed.  
In practice, this is mitigated by dimensionality reduction (e.g.\ PCA) or by
using larger or adaptive prototype sets, such as Growing Neural Gas or Dynamic
SOMs.  
Furthermore, since the SOM provides a piecewise--linear approximation to the
data manifold, the resulting trajectories follow the geometry of the map rather
than the true underlying curvature; nevertheless, the flows observed in complex
domains such as face embeddings remain stable and interpretable.

Finally, the generality of the approach extends beyond classical SOMs.  
Because the method relies only on prototype geometry and squared--distance structure, any architecture producing a set of reference vectors with
topological organisation---Growing SOMs, hierarchical maps, SOM--VAE, neural gas,
and related models---can serve as the backbone for inversion and for MUSIC
trajectories.  
This creates a flexible bridge between competitive learning and modern latent
representations, opening avenues for semantic editing, prototype-guided data
augmentation, and interactive exploration of learned manifolds.

\section{Conclusion}

This work revisits Self-Organizing Maps through the lens of distance geometry
and shows that their activation patterns contain sufficient information to
recover the original input exactly.  
By interpreting the squared distances to prototypes as a structured system of
affine constraints, we derived a linear inversion mechanism that is fully
deterministic, geometrically grounded, and requires no probabilistic modeling.
Building on this principle, we introduced MUSIC, a manifold-aware update rule
that generates smooth and interpretable semantic trajectories on the prototype
lattice.  
MUSIC combines the selective manipulation of squared distances with Tikhonov
regularization, producing stable flows within Voronoi regions and controlled
reorientations across their boundaries.

Across experiments on synthetic mixtures, handwritten digits, and facial
images, MUSIC reveals how SOM structure shapes latent dynamics.  
On a GMM, the method exhibits the expected behaviour of a piecewise-linear
flow driven by prototype geometry.  
On MNIST, it produces meaningful cross-class transitions with clear
intermediate states and quantifiable continuity.  
On Labelled Faces in the Wild, MUSIC enables attribute-specific transformations and
user-defined semantic edits while preserving identity and photometric
coherence.

The quantitative evaluation on MNIST provides concrete evidence for the advantages of topology-guided interpolation.
Across ten digit pairs, MUSIC produces significantly sharper intermediate images than linear interpolation ($p\!=\!0.002$, all pairs favouring MUSIC) and maintains higher minimum classifier confidence on the majority of transitions.
A classifier confidence analysis reveals that MUSIC transitions directly between source and target class basins, while linear interpolation detours through semantically unrelated regions.
The hyperparameter study confirms practical robustness: mean confidence varies by less than one percentage point across the full range of the preservation weight, and the Tikhonov parameter exhibits well-behaved regularisation dynamics consistent with the theory.

A key strength of the framework is its architectural generality.  
Because inversion and MUSIC rely solely on prototype coordinates and
squared--distance geometry, they extend naturally to the whole family of
prototype--based models: classical SOMs, Growing Neural Gas, Dynamic and
Hierarchical SOMs, adaptive--topology variants, and hybrid approaches such as
SOM--VAE.  
In all cases, the prototypes provide the coordinate chart while MUSIC supplies
the continuous dynamics.

The method also comes with clear limitations.  
Reconstruction fidelity is bounded by the expressiveness of the latent basis
(e.g., PCA, autoencoder embeddings), and cannot match the realism of modern
generative decoders.  
The SOM approximates the data manifold only piecewise linearly, so curvature is
captured discretely rather than continuously.  
Nonetheless, these constraints are precisely what endow the approach with
interpretability: prototype geometry makes latent transitions transparent,
stable, and semantically meaningful.

Overall, this work bridges classical competitive learning with modern latent--space analysis.
SOM inversion transforms prototype maps into deterministic coordinate charts, and MUSIC provides a mechanism for navigating them in a controlled, manifold-aware fashion.
This combination enables interpretable semantic exploration, latent editing, and data augmentation, and indicates a broader role for prototype-based representations within geometric and generative machine learning.

\clearpage

%Bibliography
% \bibliographystyle{unsrt} 
\bibliographystyle{unsrt}
\bibliography{references}  

\clearpage

%%%%%%%%%%%%%%%%%%%%%%%%%%%%%%%%%

\appendix

% --- Supplement numbering scheme ---
\renewcommand{\thesection}{S\arabic{section}}
\renewcommand{\thesubsection}{S\arabic{section}.\arabic{subsection}}

\setcounter{section}{0}
\setcounter{figure}{0}
\setcounter{table}{0}
\renewcommand{\thefigure}{S\arabic{figure}}
\renewcommand{\thetable}{S\arabic{table}}

%%%%%%%%%%%%%%%%%%%%%%%%%%%%%%%%%

% Supplementary
\section*{Supplementary Material}
\setcounter{section}{0}
\renewcommand{\thesection}{S\arabic{section}}

%%%%%%%%%%%%%%%%%%%%%%%%%%%%%%%%%%%%%%%%%%

\section{Additional properties of the SOM activation map}
\label{sec:supp_activation}

This section collects three results on the SOM squared-distance 
activation map $f(z) = a(z)$ that complement the inversion theory 
developed in Section~\ref{sec:theory} of the main text.  They are 
included here for completeness; no later derivation depends on them 
directly.

\subsection{Equivariance under rigid motions}
\label{supp:equivariance}

A useful property of the activation map is its invariance under rigid 
motions applied jointly to the input and the prototypes.  For any 
translation $t\in\mathbb{R}^D$ and orthogonal transformation 
$R\in O(D)$,
\[
a_j(Rz+t)=\|Rz+t-(Rw_j+t)\|^2=\|z-w_j\|^2 = a_j(z).
\]
Hence, the activations depend only on the \emph{relative} configuration 
of $z$ and $\{w_j\}$, not on the absolute position or orientation of 
the point cloud.  This confirms that centering or whitening the 
prototypes (as recommended in Section~\ref{sec:theory}) does not alter 
the activation pattern and therefore cannot affect the inversion result.

\subsection{Full rank of the activation Jacobian}
\label{supp:jacobian_proof}

The main text states that the Jacobian of the activation map has full 
rank everywhere (used in the perturbation analysis of 
Section~\ref{sec:PerturbingSOM}).  We provide the proof here.

\begin{lemma}[Full rank of the activation Jacobian]
\label{lemma:jacobian_rank}
Let $J(z)\in\mathbb{R}^{N\times D}$ be the Jacobian with rows 
$J_j(z)=\nabla_z a_j(z)^\top = 2(z-w_j)^\top$.  If the prototypes 
$\{w_j\}$ contain $D{+}1$ affinely independent points, then 
$\mathrm{rank}\,J(z)=D$ for all $z\in\mathbb{R}^D$.
\end{lemma}

\begin{proof}
Fix any anchor~$r$.  Since $\{w_j-w_r\}_{j\neq r}$ span 
$\mathbb{R}^D$ by assumption, the vectors 
$\{z-w_j\}_{j\neq r} = \{(z-w_r)-(w_j-w_r)\}_{j\neq r}$ span the 
same subspace.  Hence the rows of $J(z)$ span $\mathbb{R}^D$, giving 
$\mathrm{rank}\,J(z)=D$.
\end{proof}

\subsection{Weighted least-squares inversion under correlated noise}
\label{supp:weighted_ls}

Corollary~\ref{cor:stability} in the main text assumes i.i.d.\ 
activation noise.  When the noise has a known covariance structure 
$\Sigma$ (i.e., $\varepsilon\sim\mathcal{N}(0,\Sigma)$), a tighter 
reconstruction is obtained via weighted least squares:
\[
\hat z \;=\; \arg\min_z \,\|(Bz-c)\|_{\Sigma^{-1}}^2
\;=\; (B^\top \Sigma^{-1}B)^{-1}\,B^\top \Sigma^{-1}c,
\]
with stability now governed by the smallest eigenvalue of 
$B^\top \Sigma^{-1}B$.  This is relevant whenever downstream noise 
sources (e.g., quantization or sensor error) introduce heteroscedastic 
perturbations to the activation vector.

\section{Radial-tangential decomposition and global isodistance preservation}
\label{supp:S1}

Consider a fixed input $z \in \mathbb{R}^D$ and a set of SOM prototypes $\{w_j\}_{j=1}^N$. 
The squared-distance activation of prototype $j$ is
\[
a_j(z) = \|z - w_j\|^2,
\]
with first–order change under a small perturbation $\Delta z$ given by
\[
\Delta a_j \;\approx\; 2 (z - w_j)^\top \Delta z.
\]
Thus, preserving the distance to prototype $j$ to first order amounts to enforcing
\[
(z - w_j)^\top \Delta z = 0,
\]
i.e., $\Delta z$ lies in the orthogonal complement of the radial direction $z - w_j$.

Fix a reference prototype $w_k$ and define $x = z - w_k$. 
We decompose any perturbation as
\[
\Delta z = \Delta z_{\parallel} + \Delta z_{\perp},
\qquad
\Delta z_{\parallel} = \alpha x,\quad
x^\top \Delta z_{\perp} = 0.
\]
By construction, the parallel component $\Delta z_{\parallel}$ controls the first–order change of the distance to $w_k$, while the orthogonal component $\Delta z_{\perp}$ lies in the tangent subspace of the local isodistance manifold around $z$ relative to $w_k$.

To see why $\Delta z_{\perp}$ also preserves the distances to \emph{all} prototypes to first order, note that
\[
z - w_j 
= (z - w_k) + (w_k - w_j) 
= x + c_j,
\]
where $c_j = w_k - w_j$ is a constant vector independent of $\Delta z$. 
Hence the family of radial directions $\{z - w_j\}_{j=1}^N$ lies in an affine translate of the one–dimensional subspace spanned by $x$:
\[
z - w_j \in x + \mathrm{span}\{c_j : j=1,\dots,N\}.
\]
As a consequence, all first–order distance changes can be written as
\[
\Delta a_j 
\;\approx\; 
2 (z - w_j)^\top \Delta z
=
2 (x + c_j)^\top \Delta z
=
2 x^\top \Delta z + 2 c_j^\top \Delta z.
\]
If we choose $\Delta z_{\perp}$ such that $x^\top \Delta z_{\perp} = 0$, then
\[
\Delta a_j^{(\perp)} 
\;\approx\; 
2 (z - w_j)^\top \Delta z_{\perp}
=
2 x^\top \Delta z_{\perp} + 2 c_j^\top \Delta z_{\perp}
=
2 c_j^\top \Delta z_{\perp}.
\]

The vectors $c_j = w_k - w_j$ are fixed prototype offsets that do not depend on the current point $z$.  
Although their magnitudes may vary across prototypes, the corresponding contributions
$\,c_j^\top \Delta z\,$ remain uniformly bounded.  
In contrast, the shared term 
\[
x = z - w_k
\]
captures the large-scale geometry of $z$ relative to the SOM prototypes, and the component 
$\,x^\top \Delta z\,$ typically dominates the first--order variation of all squared distances.  
Thus, a perturbation $\Delta z$ decomposes into a dominant component along $x$, which controls the primary variation of $\|z - w_j\|^2$, and a residual component orthogonal to $x$, whose effect is mediated only through the bounded offsets $c_j$.  
Consequently, the contribution of $\Delta z_{\perp}$ to the squared distances is much smaller—effectively a second-order effect—compared to the variation produced by the component of $\Delta z$ parallel to $x$.

Figure~\ref{fig:perturbation_orthogonal_component} illustrates this decomposition in $\mathbb{R}^2$: the radial direction $x = z - w_k$ controls distance changes to the reference prototype, while tangential directions orthogonal to $x$ induce negligible first–order changes in the distances to all prototypes.

\paragraph{Typical dependence on the ambient dimension $D$.}
The preceding argument shows that tangential motion eliminates the shared contribution $x^\top \Delta z$ and leaves only the residual terms $c_j^\top \Delta z_\perp$.
To make the dependence on $D$ explicit, we analyze the typical magnitude of these residual dot products when $\Delta z_\perp$ is chosen without preferential alignment inside the $(D-1)$--dimensional subspace orthogonal to $x$.

Let $\mathcal{T} := \{u \in \mathbb{R}^D : x^\top u = 0,\ \|u\|=1\}$ be the unit sphere in the tangential subspace, and write $\Delta z_\perp = \eta u$ with $u \in \mathcal{T}$ and step length $\eta>0$.
Since $u \perp x$, only the component of $c_j$ orthogonal to $x$ contributes:
\[
c_{j,\perp} \;:=\; \Big(I - \frac{xx^\top}{\|x\|^2}\Big)c_j,
\qquad
c_j^\top u \;=\; c_{j,\perp}^\top u.
\]
Assume that $u$ is uniformly distributed on $\mathcal{T}$ (i.e., an isotropic direction in the tangential subspace).
By rotational invariance in the $(D-1)$--dimensional subspace, for any fixed vector $v$ with $v \perp x$ and $\|v\|=1$, the random variable $v^\top u$ has the same distribution as the first coordinate of a uniform point on $\mathbb{S}^{D-2}$.
In particular, it is mean-zero, with variance
\begin{equation}
\mathbb{E}\big[(v^\top u)^2\big] \;=\; \frac{1}{D-1}.
\label{eq:tangent_var}
\end{equation}
Therefore, for each prototype $j$,
\begin{equation}
\mathbb{E}\big[(c_j^\top u)^2\big]
\;=\;
\mathbb{E}\big[(c_{j,\perp}^\top u)^2\big]
\;=\;
\frac{\|c_{j,\perp}\|^2}{D-1},
\qquad
\Rightarrow\qquad
|c_j^\top \Delta z_\perp|
\;\text{is typically of order}\;
\eta\,\frac{\|c_{j,\perp}\|}{\sqrt{D-1}}.
\label{eq:tangent_typical}
\end{equation}
Recalling $\Delta a_j^{(\perp)} \approx 2 c_j^\top \Delta z_\perp$, this yields the typical scaling
\[
|\Delta a_j^{(\perp)}|
\;\approx\;
2\,|c_j^\top \Delta z_\perp|
\;\sim\;
2\eta\,\frac{\|c_{j,\perp}\|}{\sqrt{D-1}},
\]
showing that tangential perturbations become increasingly close to isometries as $D$ grows (in a typical-case sense).

A high--probability bound follows from standard concentration on the sphere.
For $u$ uniform on $\mathcal{T}$ and any fixed $v \perp x$ with $\|v\|=1$,
\begin{equation}
\mathbb{P}\big(|v^\top u| \ge t\big)
\;\le\;
2\exp\!\Big(-\frac{(D-2)t^2}{2}\Big).
\label{eq:sphere_tail}
\end{equation}
Applying~\eqref{eq:sphere_tail} to $v = c_{j,\perp}/\|c_{j,\perp}\|$ and a union bound over $j=1,\dots,N$ gives that, with probability at least $1-\delta$,
\begin{equation}
\max_{1\le j \le N} |c_j^\top u|
\;=\;
\max_{1\le j \le N} |c_{j,\perp}^\top u|
\;\le\;
\|c_{\perp}\|_{\max}\,
\sqrt{\frac{2\log(2N/\delta)}{D-2}},
\qquad
\|c_{\perp}\|_{\max} := \max_j \|c_{j,\perp}\|.
\label{eq:tangent_maxbound}
\end{equation}
Hence,
\begin{equation}
\max_{1\le j \le N} |\Delta a_j^{(\perp)}|
\;\approx\;
2\,\eta\,\max_j |c_j^\top u|
\;\lesssim\;
2\,\eta\,\|c_{\perp}\|_{\max}\,
\sqrt{\frac{2\log(2N/\delta)}{D-2}}.
\label{eq:tangent_maxbound_da}
\end{equation}
This bound makes explicit that, as long as $D \gg \log N$, tangential steps are simultaneously near--isometric for all prototypes with high probability.

We stress that these scalings describe typical isotropic directions in the tangential subspace; they can be weakened if the latent geometry is strongly anisotropic or if the offsets $\{c_j\}$ are systematically aligned with a low--dimensional set of semantic directions.

%%%%%%%%%%%%%%%%%%%%%%%%%%%%%%%%%%%%%%%%%%
\section{Tikhonov–regularized quadratic energy}
\label{supp:S2}

The Manifold–Aware Unified SOM Inversion and Control (MUSIC) framework defines at each iteration a perturbation $\Delta z\in\mathbb{R}^D$ of an input vector $z$ that modulates the squared distances to the prototypes $\{w_j\}_{j=1}^N$ of a Self–Organizing Map (SOM).

The activation of prototype $j$ is $a_j(z)=\|z-w_j\|^2$ with Jacobian $J_j(z)=\nabla_z a_j(z)=2\,(z-w_j)^\top$. For small displacements $\Delta z$, the first–order expansion $a_j(z+\Delta z)\approx a_j(z)+J_j(z)\Delta z$ describes the local change in activation space. 

Given a subset $S$ of prototypes whose activations should remain approximately constant (preservation set) and a subset $T$ whose activations should change toward prescribed values (target set), one defines row–normalized Jacobians $A_S\in\mathbb{R}^{|S|\times D}$ and $B_T\in\mathbb{R}^{|T|\times D}$, together with a target activation increment $b\in\mathbb{R}^{|T|}$.

The preservation constraint is $A_S\Delta z\approx 0$, while the attraction constraint is $B_T\Delta z\approx b$. Because these constraints may be redundant or inconsistent, MUSIC formulates the update as the minimizer of a Tikhonov–regularized quadratic energy
\[
E_\gamma(\Delta z)
= (1-\gamma)\|A_S\Delta z\|^2
+ \gamma\|B_T\Delta z - b\|^2
+ \lambda\|\Delta z\|^2,
\qquad \gamma\in[0,1],\;\lambda>0.
\]
The first two terms correspond to preservation and attraction, balanced by the parameter $\gamma$, while the last term introduces a Tikhonov regularizer that stabilizes the inverse problem and controls the magnitude of the perturbation. The presence of $\lambda>0$ ensures that the energy is strongly convex even when the Jacobians are rank–deficient, thus guaranteeing a unique and stable solution.

Let $H=(1-\gamma)A_S^\top A_S+\gamma B_T^\top B_T+\lambda I_D$ and $r=\gamma B_T^\top b$. The minimizer $\Delta z^\star$ satisfies $\nabla E_\gamma(\Delta z)=2(H\Delta z-r)=0$ and is therefore
\[
\Delta z^\star=H^{-1}r
=\Big((1-\gamma)A_S^\top A_S+\gamma B_T^\top B_T+\lambda I\Big)^{-1}\gamma B_T^\top b.
\]
Since $H - \lambda I$ is positive semidefinite, and $\lambda I$ is positive definite ($H\succeq\lambda I\succ0$), the solution exists, is unique, and depends smoothly on the data.

%%%%%%%%%%%%%%%%%%%%%%%%%%%%%%%%%%%%%%%%%%
\section{Tikhonov regularization}
\label{supp:S3}
The term “Tikhonov regularization” refers to the fact that $\lambda\|\Delta z\|^2$ not only penalizes the Euclidean norm of $\Delta z$ but regularizes the inversion of the composite linear operator that maps input perturbations to activation changes. Without it, the least–squares system could be ill–posed: small perturbations in the data or in the Jacobians would produce large, unstable changes in $\Delta z$. The regularizer acts as a spectral filter on the singular values of the operator.

If we stack the preservation and attraction terms into a single least–squares system,
\[
E(\Delta z) = \|M\Delta z - y\|^2 + \lambda \|\Delta z\|^2,
\qquad
M=\begin{bmatrix}\sqrt{1-\gamma}\,A_S\\[3pt]\sqrt{\gamma}\,B_T\end{bmatrix},\quad
y=\begin{bmatrix}0\\[3pt]\sqrt{\gamma}\,b\end{bmatrix},
\]
the normal equations read $(M^\top M+\lambda I)\Delta z=M^\top y$, yielding the closed–form solution
\[
\Delta z^\star=(M^\top M+\lambda I)^{-1}M^\top y.
\]
Let $M=U\Sigma V^\top$ be the singular–value decomposition (SVD) of $M$, with orthogonal matrices
$U\in\mathbb{R}^{p\times p}$, $V\in\mathbb{R}^{D\times D}$, and diagonal $\Sigma=\mathrm{diag}(\sigma_1,\ldots,\sigma_r)$
containing the non–negative singular values $\sigma_k$ of $M$. Then
$M^\top M=V\Sigma^\top\Sigma V^\top = V\,\mathrm{diag}(\sigma_k^2)\,V^\top$,
and substituting this expression into the normal equations gives
\[
(V(\Sigma^\top\Sigma+\lambda I)V^\top)\Delta z = V\Sigma^\top U^\top y.
\]
Multiplying on the left by $V^\top$ (since $V$ is orthogonal) and defining
$\tilde z = V^\top \Delta z$ decouples the system into scalar equations
$(\sigma_k^2+\lambda)\,\tilde z_k = \sigma_k(u_k^\top y)$.
Solving for each component yields
\[
\tilde z_k = \frac{\sigma_k}{\sigma_k^2+\lambda}\,\langle y,u_k\rangle,
\qquad
\Delta z^\star = V\tilde z = \sum_{k}\frac{\sigma_k}{\sigma_k^2+\lambda}\,\langle y,u_k\rangle\,v_k.
\]
Each triplet $(u_k,v_k,\sigma_k)$ corresponds to a singular mode of the linear operator $M$:
the vector $v_k$ is a direction in input space, $u_k$ its corresponding direction in activation space,
and $\sigma_k$ measures the sensitivity of activations to perturbations along $v_k$.
The factor $\sigma_k/(\sigma_k^2+\lambda)$ is the \emph{Tikhonov filter factor},
which controls how much each mode contributes to the final update.
For well–conditioned directions with $\sigma_k^2\gg\lambda$, this factor approaches $1/\sigma_k$,
so those modes are effectively inverted.
For ill–conditioned directions with $\sigma_k^2\ll\lambda$, the factor shrinks to $\sigma_k/\lambda$,
strongly suppressing unstable components.
Hence the Tikhonov regularizer acts as a \emph{spectral low–pass filter} on the Jacobian operator $M$:
it preserves well–supported modes and damps those that correspond to redundant or noisy directions
in the activation geometry.
Geometrically, the regularizer enforces smoothness by limiting the amplitude of $\Delta z$
in directions where the mapping from input to activation is uncertain or degenerate.
This explains why the Tikhonov formulation yields numerically stable and geometrically smooth
perturbations, in contrast to the unregularized least–squares inverse that would amplify small
singular values and produce erratic updates.
In the context of MUSIC, this spectral filtering plays a crucial role: since the Jacobians of SOM
activations are often highly correlated, some input directions change many activations in a
similar way, producing nearly singular modes.
The Tikhonov filter suppresses these ill–conditioned modes, ensuring that the resulting perturbation
$\Delta z^\star$ follows a coherent and stable trajectory in the latent space.
In summary, the decomposition
\[
\Delta z^\star=\sum_{k}\frac{\sigma_k}{\sigma_k^2+\lambda}\,\langle y,u_k\rangle\,v_k
\]
shows that Tikhonov regularization is not merely an $L_2$ penalty but an explicit spectral smoothing mechanism that stabilizes the inversion of the activation Jacobian and governs the smoothness of the induced dynamics in MUSIC.

The multiplicative factor $\sigma_k/(\sigma_k^2+\lambda)$ attenuates components associated with small singular values, suppressing noisy or high–frequency directions in the null space of $M$. This mechanism is the hallmark of Tikhonov regularization and explains why it produces smoother, numerically stable perturbations compared to an unregularized least–squares inverse. The operator $H$ inherits strong convexity from the $\lambda I$ term, which bounds its condition number $\kappa(H)\le \|H\|/\lambda$. As a consequence, the solution satisfies $\|\Delta z^\star\|\le \lambda^{-1}\|r\|$, and its sensitivity to data perturbations is bounded by $\|\delta\Delta z\|\le\lambda^{-1}\|\delta r\|+O(\kappa(H)\|\delta H\|)$. The parameter $\lambda$ therefore tunes the bias–variance tradeoff: increasing it improves numerical stability at the cost of smaller update amplitudes, while decreasing it recovers the least–squares solution but may amplify noise.

%%%%%%%%%%%%%%%%%%%%%%%%%%%%%%%%%%%%%%%%%%
\section{Probabilistic interpretation}
\label{supp:S4}
From a probabilistic viewpoint, the quadratic energy $E_\gamma$ can be interpreted as the negative log–posterior of a Gaussian Bayesian model that links input perturbations $\Delta z$ to desired activation changes $b$. The relation between the two is modeled as a noisy linear observation process
\[
B_T\Delta z = b + \varepsilon,
\qquad
\varepsilon \sim \mathcal{N}(0,\sigma^2 I),
\]
where $\varepsilon$ represents random uncertainty or measurement noise in the target activation constraints. This defines the likelihood
\[
p(b\,|\,\Delta z) \;\propto\; \exp\!\left[-\tfrac{1}{2\sigma^2}\|B_T\Delta z - b\|^2\right].
\]
To regularize this inverse problem, we introduce a Gaussian prior on the perturbation itself,
\[
p(\Delta z) \;\propto\; \exp\!\left[-\tfrac{1}{2\tau^2}\|\Delta z\|^2\right],
\]
which encodes the assumption that, a priori, large deviations in input space are unlikely and that updates should remain close to the origin (or to the current $z$). The posterior over $\Delta z$ given the observed activation change $b$ is then
\[
p(\Delta z\,|\,b) \;\propto\; p(b\,|\,\Delta z)\,p(\Delta z)
\;\propto\;
\exp\!\left[-\tfrac{1}{2}\!\left(
\tfrac{1}{\sigma^2}\|B_T\Delta z - b\|^2 + \tfrac{1}{\tau^2}\|\Delta z\|^2
\right)\right].
\]
Maximizing this posterior (MAP estimation) is equivalent to minimizing its negative log, yielding
\[
\Delta z^\star
= \arg\min_{\Delta z}\left(
\tfrac{1}{\sigma^2}\|B_T\Delta z - b\|^2 + \tfrac{1}{\tau^2}\|\Delta z\|^2
\right),
\]
which is precisely the Tikhonov–regularized least–squares problem with $\lambda=\sigma^2/\tau^2$. In this probabilistic reading, the regularization parameter $\lambda$ is no longer arbitrary but acquires a clear meaning as the ratio between the noise variance in the measurement model and the prior variance of admissible perturbations. Large $\lambda$ corresponds to strong confidence in the prior (small $\tau^2$), producing conservative, small–amplitude updates; small $\lambda$ corresponds to a permissive prior or highly reliable observations, allowing larger, more reactive updates.

The same reasoning extends naturally to the full MUSIC formulation, where both preservation and attraction terms appear as Gaussian likelihood components with distinct noise levels. If we denote $\varepsilon_S\!\sim\!\mathcal{N}(0,\sigma_S^2I)$ and $\varepsilon_T\!\sim\!\mathcal{N}(0,\sigma_T^2I)$ as independent noise sources for the two constraint sets, the joint likelihood reads
\[
p(y\,|\,\Delta z)
\;\propto\;
\exp\!\left[
-\tfrac{1}{2}\!\left(
\tfrac{1}{\sigma_S^2}\|A_S\Delta z\|^2
+\tfrac{1}{\sigma_T^2}\|B_T\Delta z - b\|^2
\right)\right].
\]
By identifying $\gamma=\tfrac{\sigma_S^{-2}}{\sigma_S^{-2}+\sigma_T^{-2}}$
and $\lambda=(\sigma^2/\tau^2)$ with appropriate scaling,
one recovers the composite quadratic energy $E_\gamma(\Delta z)$ of Eq.~(1).
This shows that MUSIC can be understood as a \emph{Bayesian inverse problem}
in which the tradeoff between preservation and attraction is governed by the
relative noise levels of the two constraint channels,
while the Tikhonov term plays the role of a Gaussian prior on smoothness and amplitude
of $\Delta z$.

This interpretation places MUSIC within the broader family of probabilistic inference frameworks that include ridge regression, Kalman filtering, and variational inference in Gaussian latent–variable models. In each of these, the solution represents a balance between data fidelity (likelihood) and regularity (prior), leading to minimum–variance linear estimators under Gaussian assumptions. Here, the same principle governs the flow of information between activation space and input space: the Tikhonov term ensures that the inverse mapping from activation changes to input perturbations is statistically well–posed, with $\lambda$ acting as an interpretable signal–to–noise control parameter. Consequently, MUSIC admits a principled probabilistic foundation that connects geometric regularization with Bayesian inference, providing an additional theoretical justification for its stability and smoothness properties.

%%%%%%%%%%%%%%%%%%%%%%%%%%%%%%%%%%%%%%%%%%
\section{Role of the row normalization}
\label{supp:S5}
Row normalization of $A_S$ and $B_T$ is crucial to make $\gamma$ an interpretable balance parameter: without normalization, differences in the scales of the Jacobian rows would bias the optimization toward either preservation or attraction independently of $\gamma$. Normalization equalizes the energy contribution of each prototype, ensuring that $\gamma$ consistently weights the two objectives. The explicit form of $\Delta z^\star$ also reveals the close connection with the so–called Tikhonov projector used for partial preservation. When only the preservation constraint is considered, minimizing $\|A_S\Delta z\|^2+\mu\|\Delta z\|^2$ yields the linear operator $P_\mu=I-A_S^\top(A_SA_S^\top+\mu I)^{-1}A_S$, which projects any direction onto the approximate null space of $A_S$, smoothly suppressing components that would change the preserved activations. The MUSIC update generalizes this idea by combining the projector–like preservation with the attraction term, both modulated by $\gamma$, under a single regularized inverse. In the limit $\lambda\to0$, if the combined Jacobian $(1-\gamma)A_S^\top A_S+\gamma B_T^\top B_T$ is full rank, the update converges to the exact least–squares solution; if it is rank–deficient, the Tikhonov regularizer selects the minimum–norm solution within the set of least–squares minimizers. For $\lambda\to\infty$, the update tends to zero, corresponding to complete damping. The limiting cases $\gamma=0$ and $\gamma=1$ represent pure preservation and pure attraction, respectively. Each MUSIC step can thus be interpreted as a regularized inverse mapping that converts infinitesimal activation changes into a stable and minimal–energy input displacement.

%%%%%%%%%%%%%%%%%%%%%%%%%%%%%%%%%%%%%%%%%%
\section{Free-evolution trajectories under manifold-aware and unconstrained updates}
\label{supp:S6}
To further analyze the qualitative behaviour of MUSIC, we design a free-evolution task where the system is not given any objective other than repeatedly applying random local perturbations in activation space.

Starting from the latent representation $z^{(0)}$ of an input face, at each step $t$ we randomly select a prototype $w_{j_t}$ and sample a random perturbation of its squared distance. In the MUSIC condition, this perturbation defines a linearized target constraint $B_{T}\Delta z \approx \delta$, while a small preservation set $S$ (four nearest prototypes) constrains the update to remain consistent with the local SOM geometry. We compute the MUSIC direction $d_{\mathrm{MUSIC}}$ from the Tikhonov system $(M^\top M + \lambda I)\Delta z = M^\top y$ and apply a normalized step $\Delta z^{(t)} = \eta\, d_{\mathrm{MUSIC}}/\|d_{\mathrm{MUSIC}}\|$.

In the no-constraints baseline, the update direction is simply the radial vector $(z^{(t)} - w_{j_t}) / \|z^{(t)} - w_{j_t}\|$, again rescaled to the same step length~$\eta$. Thus, the two trajectories differ only in their \emph{direction}, not in their magnitude.

Despite identical step sizes, the two evolutions diverge sharply: unconstrained updates rapidly push the latent representation into regions that no longer decode into realistic faces, producing severe geometric distortions. Conversely, MUSIC maintains consistency with the local SOM geometry and evolves the face smoothly along semantically meaningful variations. Figure~\ref{fig:free_evolution_faces} displays a typical outcome.

\section{Algorithm pseudo-code for MUSIC exploration modes}
\label{supp:algorithms}

This section provides step-by-step pseudo-code for the three MUSIC 
exploration modes defined in Section~\ref{sec:MUSIC}, including noise 
injection for enhanced coverage.

\paragraph{Noise injection.}
To enhance coverage and avoid premature alignment to local geometric 
modes, we inject small, controlled noise at three points:
(i)~\emph{step noise} in input space, adding 
$\xi_z\sim\mathcal{N}(0,\sigma_z^2 I)$ to the deterministic update 
$\Delta z$ before trust-region clipping;
(ii)~\emph{target noise} on the right-hand side, perturbing the 
desired changes as $b\leftarrow b+\xi_b$ with 
$\xi_b\sim\mathcal{N}(0,\sigma_b^2 I)$; and
(iii)~\emph{gradient jitter} by slightly perturbing the normalized 
Jacobian rows $\widehat J_j\leftarrow \widehat J_j+\Xi_j$ with small 
Frobenius norm $\|\Xi_j\|_F$.
All noises are clipped by the trust radius to preserve local validity.
In the cluster regime, \emph{random target sub-sampling} 
$T^{(i)}\subseteq T$ at each relinearization pass further promotes 
diversity.

\begin{table}[h!]
\centering
\caption{Free exploration with geometry preservation and input-space step noise.}
\begin{tabular}{p{0.97\linewidth}}
\toprule
\textbf{Input:} prototypes $\{w_j\}_{j=1}^N$, current point $z$, weights $W_S$, ridge $\lambda$, trust radius $\tau$, step-noise scale $\sigma_z\ge 0$.\\[3pt]
\textbf{Output:} updated input $z'$.\\[4pt]
1. Build $A_S = \mathrm{stack}\{\,2(z-w_j)^\top : j=1,\dots,N\,\}$ and form $C=(W_S A_S)^\top(W_S A_S)+\lambda I$.\\
2. Compute smallest-eigenvector $q_{\min}$ of $C$.\\
3. Deterministic step: $\Delta z_{\mathrm{det}}=\tau\,q_{\min}$ (or any vector in the low-eigenvalue subspace).\\
4. Sample noise: $\xi_z\sim\mathcal{N}(0,\sigma_z^2 I)$ and clip $\xi_z\leftarrow \min\{1,\ \tau/\|\xi_z\|\}\,\xi_z$.\\
5. Combine and clip: $\Delta z=\Delta z_{\mathrm{det}}+\xi_z$, then $\Delta z\leftarrow \min\{1,\ \tau/\|\Delta z\|\}\,\Delta z$.\\
6. Update $z' = z + \Delta z$ and relinearize at $z'$.\\
\bottomrule
\end{tabular}
\end{table}

\begin{table}[h!]
\centering
\caption{Informed exploration (single target $t$) with target-noise and step-noise.}
\begin{tabular}{p{0.97\linewidth}}
\toprule
\textbf{Input:} prototypes $\{w_j\}$, current $z$, target $t$, weights $W_S,W_T$, penalties $\gamma,\lambda$, trust $\tau$, fraction $\eta$, noise scales $\sigma_b,\sigma_z\ge 0$.\\[3pt]
\textbf{Output:} updated input $z'$.\\[4pt]
1. Set $T=\{t\}$, $S=\{1,\dots,N\}\setminus T$.\\
2. Build $A_S,B_T$ from normalized Jacobians $\widehat J_j=2(z-w_j)^\top/\|z-w_j\|$.\\
3. Desired change (squared distance): $b=-\eta\,a_t(z)$; add target-noise $b\leftarrow b+\xi_b$, $\xi_b\sim\mathcal{N}(0,\sigma_b^2)$.\\
4. Solve normal equations:
\[
\big[(W_SA_S)^\top(W_SA_S)+\gamma(W_TB_T)^\top(W_TB_T)+\lambda I\big]\Delta z_{\mathrm{det}}
=\gamma(W_TB_T)^\top(W_T b).
\]
5. Sample step-noise $\xi_z\sim\mathcal{N}(0,\sigma_z^2 I)$; set $\Delta z=\Delta z_{\mathrm{det}}+\xi_z$.\\
6. Trust region: if $\|\Delta z\|>\tau$, rescale $\Delta z\leftarrow \tau\,\Delta z/\|\Delta z\|$.\\
7. Update $z' = z+\Delta z$; relinearize and optionally repeat (1--3 passes).\\
\bottomrule
\end{tabular}
\end{table}

\begin{table}[h!]
\centering
\caption{Cluster exploration (multi-target): random target sub-sampling, single-prototype randomization, and noise-augmented updates.}
\begin{tabular}{p{0.97\linewidth}}
\toprule
\textbf{Input:} prototypes $\{w_j\}$, current $z$, full target set $T$, non-targets $S$, weights $W_T,W_S$, penalties $\gamma,\lambda$, trust $\tau$, passes $m$, fraction $\eta$, sub-sampling rule (size $k$ or keep-prob $p$), optional single-prototype flag, noise scales $\sigma_b,\sigma_z\ge 0$.\\[3pt]
\textbf{Output:} updated input $z'$.\\[4pt]
1.~Initialize $z^{(0)}\leftarrow z$.\\
2.~For $i=0,\dots,m-1$:\\
\hspace{0.6em}a.~\emph{Random target selection:}\\
\hspace{1.6em}--~\textbf{Multi-target sub-sampling:} draw $T^{(i)}\subseteq T$ either by fixed size $|T^{(i)}|=k$ or Bernoulli keep-probability $p$.\\
\hspace{1.6em}--~\textbf{Single-prototype randomization (optional):} draw a single target $t^{(i)}\!\sim\!\mathrm{Uniform}(T)$ and set $T^{(i)}=\{t^{(i)}\}$.\\
\hspace{0.6em}b.~Build $A_S^{(i)}, B_{T^{(i)}}^{(i)}$ from local Jacobians at $z^{(i)}$; apply weights $W_S,W_T$.\\
\hspace{0.6em}c.~Desired target changes: $b^{(i)}_t=-\eta\,a_t\big(z^{(i)}\big)$ for $t\in T^{(i)}$; add $\xi_b^{(i)}\!\sim\!\mathcal{N}(0,\sigma_b^2 I)$: $b^{(i)}\!\leftarrow\! b^{(i)}+\xi_b^{(i)}$.\\
\hspace{0.6em}d.~Solve $(A_w^\top A_w + B_w^\top B_w + \lambda I)\,\Delta z_{\mathrm{det}}^{(i)} = B_w^\top b_w$, with $A_w=W_S^{1/2}A_S^{(i)}$, $B_w=\sqrt{\gamma}\,W_T^{1/2}B_{T^{(i)}}^{(i)}$, $b_w=\sqrt{\gamma}\,W_T^{1/2}b^{(i)}$.\\
\hspace{0.6em}e.~Sample step-noise $\xi_z^{(i)}\!\sim\!\mathcal{N}(0,\sigma_z^2 I)$ and set $\Delta z^{(i)}=\Delta z_{\mathrm{det}}^{(i)}+\xi_z^{(i)}$.\\
\hspace{0.6em}f.~Trust region: $\Delta z^{(i)}\!\leftarrow\! \min\{1,\ \tau/\|\Delta z^{(i)}\|\}\,\Delta z^{(i)}$.\\
\hspace{0.6em}g.~Update and relinearize: $z^{(i+1)} = z^{(i)} + \Delta z^{(i)}$.\\
3.~Return $z' = z^{(m)}$.\\
\bottomrule
\end{tabular}
\end{table}

\end{document}